%% file: main.tex
\documentclass{article} 
\usepackage{iclr2027_conference,times}

\input{math_commands.tex}

\usepackage{hyperref}
\usepackage{algorithm}
\usepackage{algorithmic}
\usepackage{url}
\usepackage{graphicx}
\usepackage{multirow}
\usepackage{wrapfig}
\usepackage{booktabs}
\usepackage{tabularx}
\usepackage{longtable}

\usepackage{xcolor}
\usepackage{tcolorbox}
\tcbuselibrary{breakable,skins}
\iclrfinalcopy
\newcommand{\revise}[1]{{\color{blue}#1}}

\newtcolorbox{promptbox}[1]{
    enhanced,
    breakable,
    use color stack,
    width=\linewidth,
    colback=gray!4,
    colframe=gray!45,
    colbacktitle=gray!10,
    coltitle=black,
    boxrule=0.4pt,
    arc=1.5mm,
    left=8pt,
    right=8pt,
    top=5pt,
    bottom=5pt,
    toptitle=4pt,
    bottomtitle=4pt,
    before skip=6pt,
    after skip=7pt,
    fontupper=\small,
    fonttitle=\small\bfseries,
    before upper={\setlength{\parindent}{0pt}\raggedright},
    title={\revise{#1}},
    title after break={\revise{#1\space(continued)}}
}

\newenvironment{casebox}[1]{\begin{promptbox}{#1}}{\end{promptbox}}

\title{Proactive Dialogue Policy Optimization \\ via Cognitive-State Transition}

\author{
Minghui Ma\thanks{Equal contribution.} \quad
Mengqi Chen\footnotemark[1] \quad
Bin Guo\thanks{Corresponding author.} \quad
Jingqi Liu
\\
Northwestern Polytechnical University
\\
\texttt{\{2021300120,chenmengqi,1309893591\}@mail.nwpu.edu.cn}
\\
\texttt{guob@nwpu.edu.cn}
}

\begin{document}

\maketitle
\fancyhead{}

\begin{abstract}
\vspace{-4mm}
Proactive dialogue requires agents to continually adapt their policies to user feedback while progressing toward task objectives over multiple turns. To move beyond imitation learning on static datasets, recent approaches use user simulators to collect interactive data for policy optimization. However, many simulators do not explicitly model the evolution of user cognition, limiting the consistency and state dependence of feedback across turns. Moreover, representing each action only by a high-level strategy label overlooks the large utterance space and cannot distinguish alternative realizations of the same strategy.
To this end, we jointly design a \textbf{Cog}nitive User \textbf{Sim}ulator \textbf{(Cog-Sim)} and \textbf{C}ognitive-\textbf{S}tate \textbf{T}ransition--Driven \textbf{P}olicy \textbf{O}ptimization \textbf{(CSTPO)}. Cog-Sim maintains the user's cognitive and affective states and generates responses through constrained state transitions across turns, so feedback depends on both the realized utterance and the user's current state. CSTPO organizes each action as a hierarchical strategy--utterance representation: a high-level strategy label constrains utterance sampling, and utterances are optimized within each label. Sparse complete-branch sampling reuses shared dialogue prefixes and estimates separate strategy-level and utterance-level advantages, enabling fine-grained optimization at both levels. Across three tasks, Cog-Sim exhibits monotonic dose--response relationships and is preferred over prompt-based simulators for naturalness. CSTPO improves Qwen3-14B's performance to a level comparable to that of GPT-5.5-based planning methods.
\end{abstract}
\vspace{-4mm}

\section{Introduction}
\vspace{-3mm}
Proactive dialogue agents plan actions around task objectives and continually adapt their interaction policies to user feedback. In tasks such as negotiation and emotional support, agents must make a sequence of strategic decisions to move the dialogue forward. Strategy planning and optimization are therefore central to proactive dialogue research~\citep{10.1145/3715097}. Existing studies improve proactive dialogue in two directions: strategic reasoning and planner training. Prompt-based methods use explicit reasoning chains to analyze the dialogue context, derive an action plan, and generate a response. They allow language models to organize goal-directed strategies from the current context~\citep{deng-etal-2023-prompting, 2023arXiv231211792C, zhang2023ask}. However, these methods mainly use the model's existing planning ability. They do not accumulate interaction experience through parameter updates. Supervised approaches instead train planners from dialogue demonstrations with strategy, reasoning, and action-plan annotations. The annotations can be created by human experts or by automatic labeling~\citep{ito2025enhancing, 10.1145/3774904.3792335}. Such methods learn decision patterns from demonstrations, but they mainly fit existing behavior and do not repeatedly adjust the policy from feedback generated by the agent's own actions.

To address the lack of interactive trial, simulator-based reinforcement learning (RL) trains policies in dynamic dialogue environments. At each turn, the agent selects a dialogue strategy and generates an utterance. A user simulator responds based on its role, task objective, and dialogue history. The resulting task return is used to update the planner. PPDPP follows this paradigm and further trains a strategy planner through simulated interaction~\citep{deng2024plug}. The simulator thus serves as the environment for policy exploration. Its feedback shapes both the dialogue history for subsequent decisions and the resulting RL trajectory.
However, prompt-based user simulators mainly constrain responses through task context, user identity, and initial conditions. Information interpretation, stance revision, and intention formation remain implicit in language-model inference. Persistent state representations and explicit transition constraints remain largely absent~\citep{deng2024plug}. Conflating argument acceptance with action commitment risks premature goal pursuit or repeated use of mismatched strategies. Prior work also reports excessive acceptance or persistent rejection in some simulations, raising concerns about biased training feedback~\citep{hu2025astro}. Recent studies introduce richer user profiles, latent concerns, and
explicit per-turn state dynamics~\citep{Wu2026HumanLMSU, Zhang2026UserLMR1MH,zhang2026unlocking}. However, profile variables are often fixed before interaction, while goal-oriented state variables may be designed primarily to track progress toward task completion. 

\begin{wrapfigure}{r}{0.45\textwidth}
    \centering
    \vspace{-7pt}
    \includegraphics[width=\linewidth]{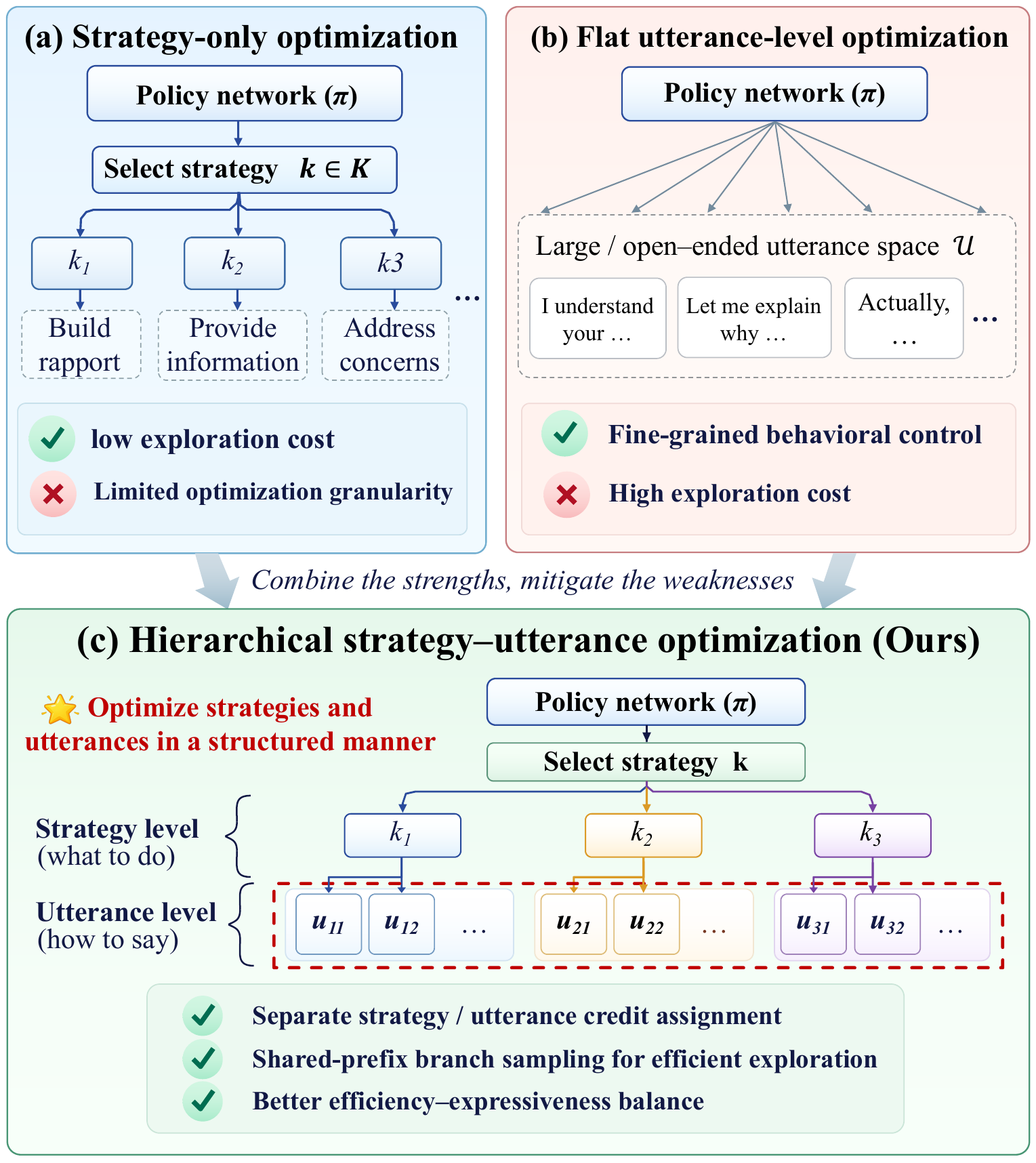}
    \vspace{-4mm}
    \caption{Three policy optimization paradigms for LLM-based proactive dialogue: (a) strategy-only, (b) flat utterance-level, and (c) hierarchical strategy--utterance optimization paradigm. \textit{CSTPO} decomposes each action into a strategy and its conditioned utterance realization.}
    \label{fig:introduction}
    \vspace{-6pt}
\end{wrapfigure}

Limited user modeling is accompanied by restricted agent-side optimization. Proactive dialogue methods typically optimize over a fixed, finite strategy set and use a language model for utterance generation. Some freeze the generator to avoid reward-driven language degradation~\citep{li2024dialogue}. Label-only actions limit optimization granularity and obscure differences between utterances under the same strategy. Existing methods use paired rollouts or forward sampling to inform current decisions through future interaction outcomes~\citep{wu2025collabllm,ICLR2025_186094ef}. Others improve value estimation through explicit search or privileged training information~\citep{he2024planning,zhang2026unlocking}. However, terminal-only rewards still require complete trajectories for comparison, making trajectory-level sampling a major efficiency bottleneck. General multi-turn RL frameworks support richer response-level optimization by combining higher-level value estimation with token-level policy updates~\citep{zhou2024archer}. However, their high-level value functions operate at the response level rather than over explicit strategy variables. Thus, restricted action spaces control sampling costs but limit behavioral granularity, whereas richer spaces require more trajectory comparisons.

These observations lead to two challenges. \textit{\textbf{First, user feedback cannot be reliably attributed to cognitive change.}} Without persistent state records and transition constraints, changes in information acceptance and action intention remain implicit. Simulators may thus miss state-dependent responses to the same utterance. \textit{\textbf{Second, action expressiveness and sampling efficiency are difficult to balance.}} Richer action spaces require more comparisons, but terminal rewards demand complete interactions for evaluation. The interaction budget does not scale with these comparisons.

To address these challenges, we propose a proactive dialogue policy optimization framework integrating \textit{Cog-Sim} and \textit{CSTPO}. \textit{\textbf{Cog-Sim}} maintains a dynamic cognitive–affective state and generates feedback through constrained cross-turn state transitions.
Its explicitly maintained internal states allow multiple counterfactual continuations to be sampled from the same pre-action state.
As shown in Figure~\ref{fig:introduction}, \textit{\textbf{CSTPO}} represents actions through a strategy–utterance hierarchy. Strategy labels constrain utterance generation, while comparisons among utterances under the same strategy refine how that strategy is expressed. During training, the value model jointly estimates state and strategy-conditioned values to derive separate strategy- and utterance-level advantages from terminal rewards. 
We evaluate the framework on three multi-turn dialogue tasks: emotional support, donation persuasion, and negotiation. Cog-Sim exhibits consistent dose--response behavior under increasing intervention strength and is preferred over prompt-based simulators for naturalness. CSTPO consistently outperforms prompting and learning-based baselines on task outcomes. With a smaller Qwen3-14B backbone, it achieves performance comparable to GPT-5.5-based planners. Ablations show that cognitive-state modeling improves policy optimization, while the hierarchical Critic enables fine-grained credit assignment between strategy selection and utterance realization.

\section{Related Work}
\paragraph{Policy planning for proactive dialogue}
\label{sec:related-policy-planning}

Proactive dialogue policy planning aims to optimize long-term dialogue outcomes through sequential decisions. Prompt-based methods such as Proactive and ProCoT~\citep{deng-etal-2023-prompting}, Ask an Expert~\citep{zhang2023ask}, and ICL-AIF~\citep{fu2023improving} elicit strategic behavior through strategy selection, explicit reasoning, expert advice, or interaction feedback, but do not directly optimize policy parameters for long-term returns. Supervised approaches instead learn parameterized planners from annotated demonstrations; for example, \citet{ito2025enhancing} fine-tune ProCoT with automatically annotated reasoning traces and action plans. More recent methods move beyond imitation by optimizing policies through interaction. PPDPP~\citep{deng2024plug} combines supervised initialization with policy-gradient training, while DialogXpert~\citep{rakib2026dialogxpert} uses Deep Q-learning with emotion trajectories and LLM action priors. PRINCIPLES~\citep{kim-etal-2025-principles} constructs reusable strategy memory through offline self-play, and LDPP~\citep{he2025simulation} learns latent strategy representations with offline hierarchical RL. These methods improve strategic decision making, but policy expressiveness remains tied to either predefined strategy sets or learned strategy abstractions. CSTPO instead explicitly factorizes each turn into a strategy and its conditioned utterance, allowing both levels to be optimized from interaction outcomes.

\paragraph{User simulation in dialogue policy learning}

User simulators provide the interaction environment for dialogue policy learning, and simulator design directly affects downstream policy quality and transfer~\citep{suh2026quantifying}. Classical goal-driven simulators maintain task consistency through explicit agendas or learned behavior models~\citep{schatzmann2007agenda,lin2022gentus}; Deep Dyna-Q further showed that simulator bias can directly affect learned dialogue policies~\citep{peng-etal-2018-deep}. Profile-conditioned simulators model user heterogeneity and persona-consistent behavior~\citep{zhao-etal-2024-esc,zhang-etal-2024-strength,wang2025know,abdulhai2026consistently}, but such profiles are typically fixed before interaction and do not represent how a user's beliefs, concerns, or intentions change in response to the agent. State-based approaches make internal user conditions or cognitive trajectories explicit for generation and evaluation~\citep{Wu2026HumanLMSU,ma2026cognitive}, and recent work further incorporates dynamic states into policy learning, reward design, and credit assignment~\citep{Zhang2026UserLMR1MH,kim2026discoverllm,zhang2026unlocking}. However, many formulations emphasize progress toward task completion rather than controlled bidirectional changes in user cognition. Cog-Sim models cognitive and affective states as persistent variables whose transitions can move in either direction under theory-grounded, parameterized constraints, enabling state-dependent and counterfactual user responses.

\paragraph{Multi-turn RL and credit assignment}
Multi-turn dialogue and language-agent learning must assign delayed outcome rewards to intermediate decisions. ArCHer~\citep{zhou2024archer} addresses this problem with hierarchical RL, using utterance-level value estimation to guide token-level policy optimization. REFUEL~\citep{ICLR2025_186094ef} instead estimates relative future values from self-generated interactions to improve multi-turn policy optimization. More recently, Tree Search for LLM Agent Reinforcement Learning~\citep{ICLR2026_8c7304e7} organizes rollouts into shared-prefix trees, reducing redundant interaction while deriving finer-grained learning signals from terminal outcomes. Dialogue-specific work also explores hierarchical strategy and response optimization. \citet{zhao-etal-2025-chain} construct turn-level strategy--response preference pairs through Monte Carlo Tree Search for emotional support, while \citet{lin-etal-2026-dual} coordinate strategic dialogue management and fine-grained utterance generation with dual hierarchical policies. These methods improve credit assignment across turns or between decision levels, but they do not explicitly decompose the value of a dialogue action into the contribution of strategy selection and that of its linguistic realization. CSTPO introduces this decomposition through strategy-conditioned value estimation and shared-prefix branching, yielding separate inter-strategy and intra-strategy advantages from terminal rewards.

\input{sections/method}

\input{sections/setup}

\input{sections/experiment}



\section{Conclusion}
\vspace{-3mm}

In this work, we connect user cognition with proactive dialogue policy learning through Cog-Sim and CSTPO. Cog-Sim grounds user responses in explicit cognitive and affective states, while CSTPO enables agents to jointly optimize strategic decisions and linguistic realizations. Experiments across emotional support, donation persuasion, and negotiation show that Cog-Sim provides more consistent and natural user feedback than prompt-based simulators, and CSTPO-trained Qwen3-14B achieves higher terminal returns than strong baselines including GPT-5.5-based ProCoT and PPDPP. Reliable user simulation remains an important challenge for interactive policy learning. Future work will calibrate cognitive and affective transitions with human interaction data and investigate broader user profiles and contexts. More comprehensive evaluation combining independent evaluators, cross-simulator validation, and human interaction studies will be important for understanding whether simulated improvements translate to effective real-world dialogue.



\clearpage
\section*{AI Use Statement}
Generative AI tools were used at several stages of this work. \texttt{Codex}%
\footnote{\url{https://openai.com/codex/}} and \texttt{Claude Code}%
\footnote{\url{https://www.anthropic.com/claude-code}} were used to assist with conceptual discussions of \textit{Cog-Sim} and \textit{CSTPO}, the initial design of the experimental protocol, implementation of experimental code, hyperparameter selection, and analysis of experimental results. The authors revised the proposed experimental procedures and hyperparameters throughout the project to reflect the actual experimental settings and observations.

\texttt{GPT-5.5} was used to generate teacher trajectories for supervised training. LLMs were also used for strategy annotation: each instance was annotated independently three times, with the majority strategy retained when available; cases without agreement were manually annotated by the authors. AI tools were not used for mathematical derivations or proofs.

Generative AI was additionally used to assist with drafting and language polishing to improve clarity and fluency without changing the intended technical meaning. Literature retrieval, reference selection, and \texttt{BibTeX} organization were performed by the authors. All AI-assisted code, generated trajectories, annotations, experimental analyses, and manuscript text were repeatedly reviewed and verified by three authors.

\subsection*{Ethics Statement}

All datasets used in this work are publicly available:
ESConv\footnote{\url{https://github.com/thu-coai/Emotional-Support-Conversation}},
PersuasionForGood\footnote{\url{https://gitlab.com/ucdavisnlp/persuasionforgood}},
and CraigslistBargain\footnote{\url{https://github.com/stanfordnlp/cocoa/tree/master/craigslistbargain}}.
We do not collect new conversational data from real end users. Human evaluation is conducted by five graduate students with backgrounds in natural language processing. Annotators perform blind pairwise comparisons, participate with informed consent, and receive reasonable compensation according to local standards.

Our cognitive and affective state representations are designed solely for dialogue simulation and policy learning. They should not be interpreted as clinical assessments, psychological diagnoses, or reliable estimates of the internal states of real users. Similarly, although we study persuasion as a benchmark task, optimizing persuasive dialogue policies may raise concerns about undue influence if such systems are deployed with real users. We therefore evaluate the proposed methods only on public datasets and simulated interactions and do not claim readiness for real-world deployment.

We do not evaluate demographic fairness because the three public datasets do not provide a consistent set of demographic attributes. This remains a limitation, as both user simulators and learned policies may inherit biases from the underlying datasets and language models.



\clearpage
\bibliography{iclr2027_conference}
\bibliographystyle{iclr2027_conference}

\appendix
\input{sections/appendix}

\end{document}

%% file: math_commands.tex
\usepackage{amsmath,amsfonts,bm}

\def\eqref#1{equation~\ref{#1}}

\def\1{\bm{1}}

\DeclareMathAlphabet{\mathsfit}{\encodingdefault}{\sfdefault}{m}{sl}
\SetMathAlphabet{\mathsfit}{bold}{\encodingdefault}{\sfdefault}{bx}{n}



%% file: sections/method.tex
\section{Method}
\label{sec:method}

\begin{figure}[t]
    \centering
    \includegraphics[width=\linewidth]{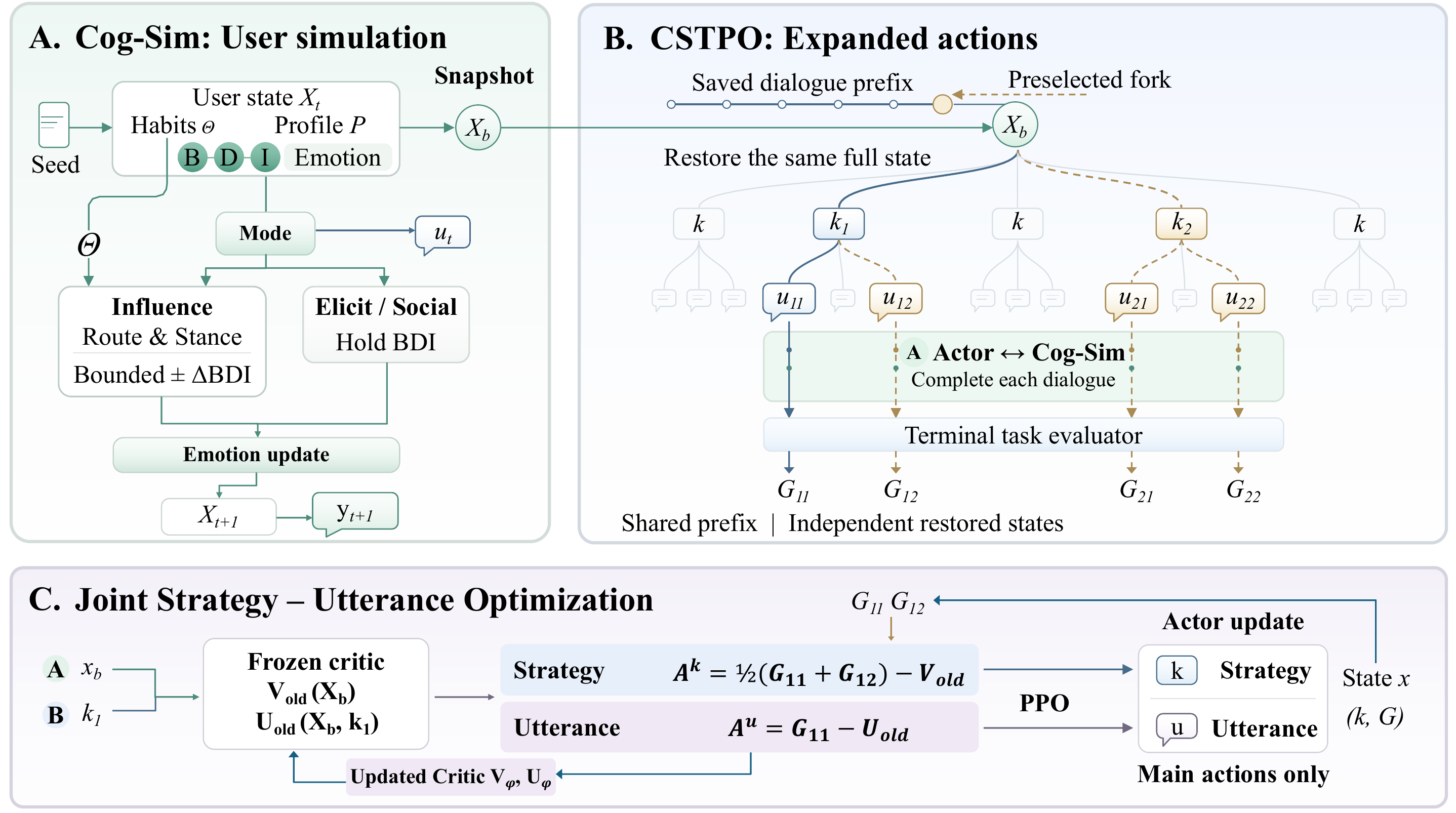}
    \vspace{-4mm}
    \caption{Overview of the proposed framework. (a) Cog-Sim simulates cognitive-affective user state transitions; (b) CSTPO performs shared-prefix rollout with hierarchical strategy--utterance actions; (c) CSTPO optimizes strategy selection and utterance realization with separate credit assignment.}
    \label{fig:framework}
    \vspace{-2mm}
\end{figure}

\subsection{Problem Formulation}\
We model proactive dialogue as a finite-horizon, partially observed interaction with sparse terminal feedback. An episode starts from an interaction seed $z\sim\mathcal D$. At turn $t$, $x_t$ denotes the training-side pre-action context. It includes the dialogue history, simulated user profile and state, task variables, and remaining horizon. The Actor observes only $o_t=\mathcal O(x_t)$, which contains the dialogue history and role-visible task information.
At each turn, the Actor takes a hierarchical action $a_t=(k_t,u_t)$. This action consists of a strategy label $k_t\in\mathcal K$ and an utterance $u_t$ conditioned on the selected strategy. The policy therefore factorizes as follows:

\begin{equation}
\pi_\theta(a_t\mid o_t)
=
\pi_\theta(k_t\mid o_t)
\pi_\theta(u_t\mid o_t,k_t).
\label{eq:policy_factorization}
\end{equation}

The utterance $u_t$ is then passed to Cog-Sim, which generates the next user observation. The resulting interaction trajectory is $\tau=(o_1,a_1,\ldots,o_T,a_T,o_{T+1})$.
After the dialogue ends, a terminal task utility $G(\tau)$ is assigned. The learning objective is

\begin{equation}
J(\theta)
=
\mathbb E_{z\sim\mathcal D,\,
\tau\sim p_\theta(\cdot\mid z)}
\left[G(\tau)\right],
\label{eq:objective}
\end{equation}

where $p_\theta(\tau\mid z)$ denotes the trajectory distribution induced by the Actor and Cog-Sim.

\subsection{Cog-Sim: Cognitive-State-Driven User Simulation}

Cog-Sim represents each user with a persistent user profile $P$ and a dynamic cognitive-affective state $s_t^{u}=(C_t,E_t)$. Specifically, $C_t=(B_t,D_t,I_t)$ denotes beliefs, desires, and intentions under the BDI framework~\cite{rao1995bdi}, while $E_t=(v_t,\rho_t,c_t)$ denotes affective valence, arousal, and a discrete emotion category~\cite{smith1985patterns,russell1980circumplex}. A user-specific parameter vector $\Theta_u$ captures stable cognitive processing tendencies. Parameterized by $\Theta_u$, $\operatorname{Update}_{\Theta_u}$ uses the Elaboration Likelihood Model and Social Judgment Theory~\cite{petty1986elaboration,sherif1961social} to infer the user's processing route and stance, then applies deterministic constraints to an LLM-proposed BDI change.

Given the dialogue history $H_t$ and the realized Actor utterance $u_t$, a task-agnostic Action Type Classifier (ATC) assigns $q_t=\operatorname{ATC}(H_t,u_t)\in\{\mathrm{Influence},\mathrm{Elicit},\mathrm{Social}\}$. Influence denotes an attempt to change the user's cognition, Elicit requests existing information, and Social covers affective or relational interaction. Unlike the task-specific strategy $k_t$, $q_t$ is inferred from the actual utterance and controls the simulator transition. Influence invokes the cognitive updater, whereas Elicit and Social preserve the BDI state; all modes subsequently update affect and generate a user response:

\begin{equation}
\begin{aligned}
C_{t+1}
&=
\begin{cases}
\operatorname{Update}_{\Theta_u}(P,C_t,E_t,H_t,u_t),
& q_t=\mathrm{Influence},\\
C_t,
& q_t\in\{\mathrm{Elicit},\mathrm{Social}\},
\end{cases}\\
(E_{t+1},y_t^{\mathrm{usr}})
&=
\operatorname{Respond}(P,C_{t+1},E_t,H_t,u_t,q_t).
\end{aligned}
\label{eq:cogsim_transition}
\end{equation}


The environment then appends $y_t^{\mathrm{usr}}$ to the dialogue history and constructs $x_{t+1}$ and $o_{t+1}=\mathcal O(x_{t+1})$. Cog-Sim stores a recoverable pre-action checkpoint at each turn. Therefore, different candidate actions can independently continue from the same user state and dialogue prefix. Appendix~\ref{app:cogsim} provides the full transition and checkpoint specifications.

\subsection{CSTPO: Cognitive-State-Assisted Hierarchical Optimization}

\textbf{Cog-Critic: Dual-value credit decomposition.} CSTPO uses a Critic with two value heads:

\begin{equation}
V^\pi(x_t)
=
\mathbb E_\pi[G\mid x_t],
\qquad
U^\pi(x_t,k_t)
=
\mathbb E_\pi[G\mid x_t,k_t].
\label{eq:dual_value}
\end{equation}

$V^\pi(x_t)$ estimates the expected terminal return before a strategy is selected. $U^\pi(x_t,k_t)$ estimates the return after selecting $k_t$ but before committing to a particular utterance. Both heads use the pre-action context, $U^\pi$ additionally accesses the strategy label.
Let $Q^\pi(x,k,u)=\mathbb E_\pi[G\mid x,k,u]$ denote the expected return after both action fields are fixed. The two heads induce the following decomposition:
\begin{equation}
Q^\pi(x,k,u)-V^\pi(x)
=
\underbrace{U^\pi(x,k)-V^\pi(x)}_{\text{strategy selection}}
+
\underbrace{Q^\pi(x,k,u)-U^\pi(x,k)}_{\text{utterance realization}}.
\label{eq:adv_decomposition}
\end{equation}
Thus, the strategy field is evaluated against the state baseline $V^\pi$, while the utterance field is evaluated against the strategy-conditioned baseline $U^\pi$.  Appendix~\ref{app:theory} formalizes state refinement, hierarchical credit decomposition, and action aliasing.

\textbf{Same-state continuations.} A single trajectory confounds the effect of strategy choice with those of wording and subsequent sampling noise. CSTPO therefore selects a subset of non-terminal checkpoints before observing the current action or its outcome. At each selected parent $b$, it restores the same pre-action context $x_b$, samples $m$ distinct strategies $\{k_{b,i}\}_{i=1}^{m}$ under the frozen old policy, and generates $\ell$ utterance realizations $\{u_{b,i,r}\}_{r=1}^{\ell}$ for each strategy. The main-path action is indexed by $(i,r)=(1,1)$; hence only $m\ell-1$ additional tails must be simulated.
Let $G_{b,i,r}$ be the terminal return obtained by continuing from $x_b$ with $(k_{b,i},u_{b,i,r})$. The strategy-level return averages over utterance realizations:
\begin{equation}
\overline G_{b,i}
=
\frac{1}{\ell}\sum_{r=1}^{\ell}G_{b,i,r}.
\label{eq:conditional_return}
\end{equation}
At an unexpanded main-path node, we set $\overline G_{t,1}=G(\tau)$. For the main action at an expanded parent, the field-level advantages are
\begin{equation}
\widehat A_b^{\,k}
=
\overline G_{b,1}-V_{\mathrm{old}}(x_b),
\qquad
\widehat A_b^{\,u}
=
G_{b,1,1}-U_{\mathrm{old}}(x_b,k_{b,1}).
\label{eq:field_advantages}
\end{equation}

Averaging returns within the main strategy reduces the dependence of its strategy advantage on a single wording realization, while returns under the alternative strategies supervise their corresponding $U(x_b,k)$ values. All additional tails are used for Critic fitting and strategy-level return estimation only; the Actor is updated exclusively with main-path actions.

\textbf{Field-aware PPO.} We distinguish strategy and utterance tokens with separate field masks. For an Actor token $w_j$ from turn $t$, its advantage is

\begin{equation}
\widehat A_j=
\begin{cases}
\widehat A_t^{\,k},
& w_j\text{ belongs to the strategy field},\\
\widehat A_t^{\,u},
& w_j\text{ belongs to the utterance field}.
\end{cases}
\label{eq:token_advantage}
\end{equation}



Let $\xi_j$ denote the context preceding token $w_j$, and let $r_j(\theta)=p_\theta(w_j\mid\xi_j)/p_{\mathrm{old}}(w_j\mid\xi_j)$ be the token-level PPO likelihood ratio. The Actor minimizes

\begin{equation}
\mathcal L_{\mathrm{Actor}}(\theta)
=
-\frac{1}{|\mathcal M|}
\sum_{j\in\mathcal M}
\min\!\left(
r_j(\theta)\widehat A_j,\,
\operatorname{clip}
\left(r_j(\theta),1-\epsilon,1+\epsilon\right)
\widehat A_j
\right),
\label{eq:actor_loss}
\end{equation}

\vspace{-2mm}

where $\mathcal M$ contains action tokens and the SFT policy initializes $\pi_\theta$. Appendix~\ref{app:cstpo} provides the full algorithm and Critic objectives.



\textbf{Terminal evaluation.}
For each completed dialogue, a frozen LLM evaluator produces a structured task judgment $s_\tau=\operatorname{Judge}(H_{T+1},c_{\mathrm{task}})$. A task-specific deterministic function then computes $G(\tau)=f_{\mathrm{task}}(s_\tau,c_{\mathrm{task}})$, where $c_{\mathrm{task}}$ contains only the whitelisted case information required for outcome evaluation. The evaluator does not access the latent cognitive state or Critic values.

\vspace{-2mm}

%% file: sections/setup.tex
\section{Experimental Setup}
\label{sec:setup}
\vspace{-2mm}

\subsection{Tasks and Evaluation}
\label{sec:setup:data}
\vspace{-2mm}

We evaluate on three multi-turn proactive dialogue tasks: \textbf{ESConv} for emotional support, \textbf{PersuasionForGood (P4G)} for donation persuasion, and \textbf{CraigslistBargain (CB)} for price negotiation. The Actor serves as the supporter, persuader, and buyer, respectively, while Cog-Sim simulates the corresponding user role. For each task, we sample 200/50/100 interaction seeds for policy training, development, and final evaluation. Dataset statistics, strategy annotations, and interaction-seed construction are provided in Appendix~\ref{app:data-action-spaces}.


The mean task-specific terminal return \textbf{$G$} is our primary metric. For ESConv, the terminal return is $G_{\mathrm{ES}}=(E+A)/8$, where Emotion Improvement (\textbf{$E$}) and Action-Plan Feasibility (\textbf{$A$}) are rated on 0--4 scales. For P4G, $G_{\mathrm{P4G}}\in\{0,1\}$ indicates a voluntary donation commitment that remains valid at dialogue termination, whereas Commitment Rate (\textbf{Comm.}) reports the proportion of dialogues containing any donation commitment. For CraigslistBargain, \textbf{Deal} reports the proportion of valid agreements, and $G_{\mathrm{CB}}=\operatorname{clip}(SL,0,1)$, where $SL=(p-p_s)/(p_b-p_s)$ is the unclipped price utility for a valid deal with final price $p$. Here, $p_b$ and $p_s$ denote the buyer's and seller's [target/reservation] prices, respectively, and higher $SL$ indicates a more favorable outcome for the buyer. We set $SL=0$ when no valid deal price can be extracted and additionally report raw-\textit{SL} before clipping. 




\subsection{Baselines}
\label{sec:setup:baseline}

\textbf{User simulators.}
We compare Cog-Sim with three prompt-based simulators: \textit{Persona}~\cite{zhao-etal-2024-esc}, which conditions on a detailed user profile; \textit{Persona-Resist}~\cite{zhang-etal-2024-strength}, which additionally discourages ungrounded agreement; and \textit{BDI-Prompt}, which exposes textual beliefs, desires, and intentions without an explicit transition mechanism.

\textbf{Dialogue policies.}
We compare against \textit{Standard} direct generation; \textit{Proactive} and \textit{ProCoT}~\citep{deng-etal-2023-prompting}, which respectively select a strategy before responding and first analyze dialogue progress to form a goal; \textit{Ask-an-Expert}~\citep{zhang2023ask}, which conditions on an expert recommendation; and \textit{MI-Prompt}~\citep{chen-etal-2023-controllable}, which converts a strategy label into a natural-language instruction. We further include two RL-based baselines: \textit{PPDPP}~\citep{deng2024plug}, a RoBERTa-large strategy planner, and \textit{DialogXpert}~\citep{rakib2026dialogxpert}, which uses an LLM prior to propose candidate actions and a learned Q-network~\citet{Mnih2013PlayingAW} to select among them. \textbf{\textit{SFT Init.}} is our supervised policy initialization without RL, while \textbf{\textit{CSTPO}} is the RL-optimized policy.

\vspace{-2mm}


\subsection{Implementation Details}
\label{sec:setup:implementation}
\vspace{-2mm}

We evaluate the prompt and planning baselines with \textbf{GPT} (\texttt{GPT-5.5}). Cog-Sim, baseline simulators, and the terminal evaluator use the \textbf{DeepSeek} (\texttt{DeepSeek-V4-Pro-0813}) backend. \textbf{Qwen} (\texttt{Qwen3-14B}) serves as the trainable backbone. Task-specific SFT policies mix human dialogues and teacher trajectories at a 1:1 ratio and use \textit{LoRA}~\citep{Hu2021LoRALA} for four epochs. RL initializes from \textit{SFT Init.}. The Actor and Critic learning rates are $1\times10^{-5}$ and $1\times10^{-4}$, respectively, and the PPO clipping coefficient is $0.2$. Each RL step samples 10 seeds. We expand 2 non-terminal checkpoints per main trajectory for P4G and CB and 4 for ESConv. At each checkpoint, we sample $m=2$ strategies and $\ell=2$ utterance realizations per strategy, Training is built on \texttt{verl}\footnote{\url{https://github.com/volcengine/verl}} with \texttt{vLLM}\footnote{\url{https://github.com/vllm-project/vllm}} and thinking disabled. Full configurations and prompts are provided in Appendices~\ref{app:implementation} and~\ref{app:prompts}.
\vspace{-2mm}

%% file: sections/experiment.tex
\section{Experiments}
\label{experiment}
\vspace{-2mm}

\subsection{User Simulator Validation}
\label{experiment:CogSim}
\vspace{-2mm}


To evaluate \textit{Cog-Sim}, we conduct a matched dose--response test against three prompt-based controls on 100 seeds per task. From identical pre-action checkpoints, each simulator responds to four input levels with two paraphrases per level. ESConv and P4G vary from weak evidence to relevant evidence and actionable proposals. For CB, levels are defined by the seller's latest asking price: L0 asks without an offer, L1 offers 95\%, L2 offers 98--99\%, and L3 accepts the full price with immediate payment. An LLM judge scores user change or concession on a 0--3 scale.

\begin{wrapfigure}{r}{0.52\columnwidth}
    \centering
    \vspace{-2mm}
    \includegraphics[width=\linewidth]{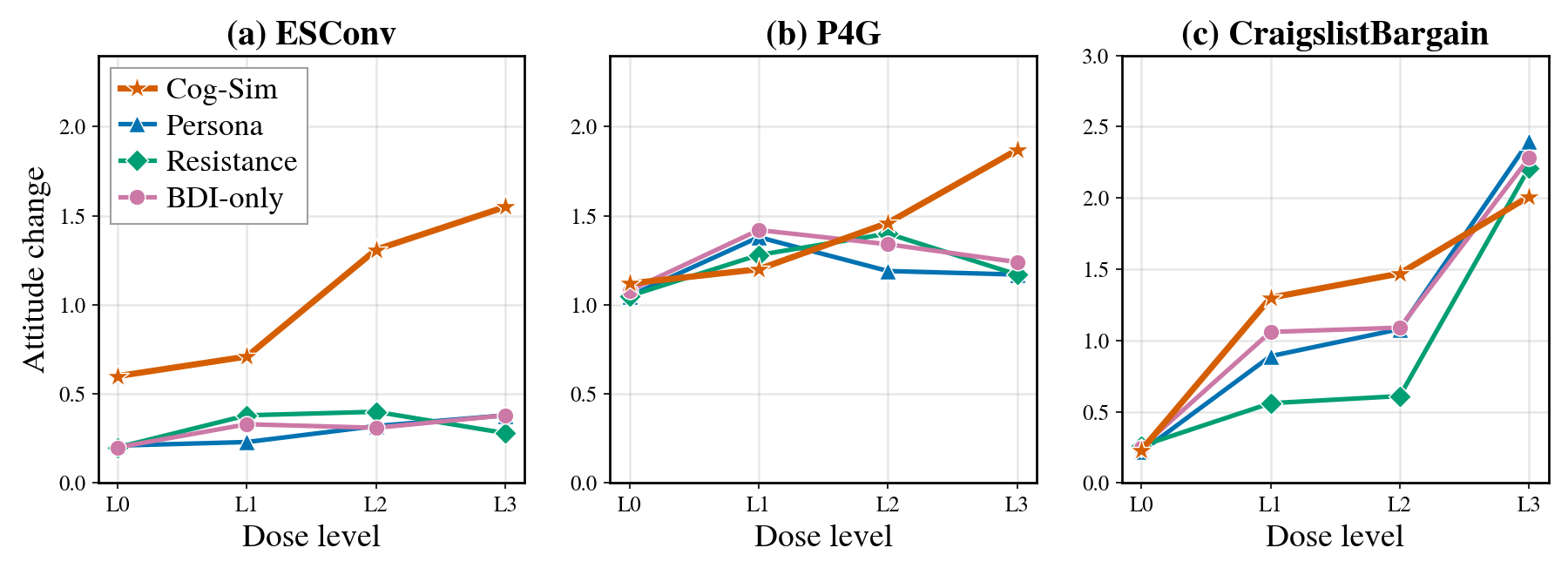}
    \vspace{-4mm}
    \caption{User responses to controlled input levels.}
    \label{fig:sim-response}
    \vspace{-2mm}
\end{wrapfigure}

Figure~\ref{fig:sim-response} shows that \textit{Cog-Sim}'s mean response scores increase monotonically on all three tasks. The scores rise from 0.60 to 1.55, from 1.12 to 1.87, and from 0.23 to 2.01, respectively. Prompt-based controls show smaller endpoint changes on ESConv and P4G. On CB, \textit{Cog-Sim} responds progressively from the intermediate offer levels, whereas the controls change most sharply only at L3; consequently, \textit{Cog-Sim} does not have the largest endpoint difference on this task.

To assess response quality, we further perform randomized pairwise evaluation of naturalness \textbf{(Nat.)},  consistency \textbf{(Cons.)}, and dose appropriateness \textbf{(App.)}, a LLM judge conducts 7,200 pairwise comparisons with randomized presentation order and hidden model identities. We report net preference as $(W-L)/N$, where $N$ counts all comparisons, including ties.

\begin{table*}[ht]
    \centering 
    \vspace{-1mm}
    \small
    \setlength{\tabcolsep}{1.0pt}
    \renewcommand{\arraystretch}{0.60}
    \begin{tabularx}{0.85\textwidth}{
        l
        *{9}{>{\centering\arraybackslash}X}
    }
        \toprule
        &
        \multicolumn{3}{c}{\textbf{ESConv}}
        & \multicolumn{3}{c}{\textbf{P4G}}
        & \multicolumn{3}{c}{\textbf{CB}} \\
        \cmidrule(lr){2-4}
        \cmidrule(lr){5-7}
        \cmidrule(lr){8-10}

        \textbf{Comparator}
        & Nat. & Cons. & App.
        & Nat. & Cons. & App.
        & Nat. & Cons. & App. \\
        \midrule

        Persona~\cite{zhao-etal-2024-esc}
        & \textbf{+.66}\textsuperscript{\ensuremath{\dagger}}
        & \textbf{+.55}\textsuperscript{\ensuremath{\dagger}}
        & \textbf{+.53}\textsuperscript{\ensuremath{\dagger}}
        & \textbf{+.63}\textsuperscript{\ensuremath{\dagger}}
        & \textbf{+.26}\textsuperscript{\ensuremath{\dagger}}
        & \textbf{+.27}\textsuperscript{\ensuremath{\dagger}}
        & \textbf{+.33}\textsuperscript{\ensuremath{\dagger}}
        & -.01
        & -.01 \\  

        Resistance~\cite{zhang-etal-2024-strength}
        & \textbf{+.41}\textsuperscript{\ensuremath{\dagger}}
        & \textbf{+.21}\textsuperscript{\ensuremath{\dagger}}
        & \textbf{+.23}\textsuperscript{\ensuremath{\dagger}}
        & \textbf{+.39}\textsuperscript{\ensuremath{\dagger}}
        & +.10
        & +.10
        & \textbf{+.42}\textsuperscript{\ensuremath{\dagger}}
        & -.09
        & +.04 \\      

        BDI-only
        & \textbf{+.60}\textsuperscript{\ensuremath{\dagger}}
        & \textbf{+.47}\textsuperscript{\ensuremath{\dagger}}
        & \textbf{+.47}\textsuperscript{\ensuremath{\dagger}}
        & \textbf{+.60}\textsuperscript{\ensuremath{\dagger}}
        & \textbf{+.23}\textsuperscript{\ensuremath{\dagger}}
        & \textbf{+.25}\textsuperscript{\ensuremath{\dagger}}
        & \textbf{+.47}\textsuperscript{\ensuremath{\dagger}}
        & +.03
        & +.05 \\
        \bottomrule
    \end{tabularx}
    \vspace{-2mm}
    \caption{
        Pairwise LLM evaluation of Cog-Sim.
        Positive values favor Cog-Sim; \ensuremath{\dagger} indicates a 95\% case-cluster bootstrap interval entirely above zero.
    }
    \label{tab:sim-preference}
\end{table*}

\textit{Cog-Sim} achieves higher naturalness preference in all nine task--comparator pairs (Table~\ref{tab:sim-preference}), indicating that explicit cognitive-state modeling improves the realism of simulated user responses. It further obtains consistent gains across all three evaluation dimensions on ESConv and shows improvements in P4G and CB, particularly in naturalness and dose appropriateness. These results demonstrate that \textit{Cog-Sim} enables more coherent and controllable user simulation across diverse interaction settings.

\subsection{Dialogue Policy Performance}
\label{experiment:main}
\vspace{-2mm}

\begin{table*}[ht]
    \centering
    \small
    \setlength{\tabcolsep}{4.3pt}

    \begin{tabular}{@{}lccccccccc@{}}
        \toprule
        &
        & \multicolumn{3}{c}{\textbf{ESConv}}
        & \multicolumn{2}{c}{\textbf{P4G}}
        & \multicolumn{3}{c}{\textbf{CB}} \\
        \cmidrule(lr){3-5}
        \cmidrule(lr){6-7}
        \cmidrule(lr){8-10}

        \textbf{Method}
        & \textbf{Backbone}
        & $E$ & $A$ & $G$
        & Comm. & $G$
        & SL & Deal & $G$ \\
        \midrule

        Standard
        & GPT
        & 2.180 & 2.627 & .601
        & \underline{45\%} & \underline{.450}
        & \underline{.571} & \underline{66\%} & .484 \\

        Proactive~\cite{deng-etal-2023-prompting}
        & GPT
        & \underline{2.487} & 2.250 & .592
        & 43\% & .430
        & .511 & 65\% & .457 \\

        ProCoT~\citep{deng-etal-2023-prompting}
        & GPT
        & \textbf{2.490} & 2.430 & .615
        & 44\% & .440
        & .509 & 61\% & .435 \\

        AnE~\citep{zhang2023ask}
        & GPT
        & 1.963 & \textbf{3.093} & \textbf{.632}
        & 42\% & .420
        & .516 & 61\% & .433 \\

        MI-Prompt~\citep{chen-etal-2023-controllable}
        & GPT
        & 2.163 & 1.620 & .473
        & 40\% & .400
        & .541 & 62\% & .456 \\

        PPDPP~\citep{deng2024plug}
        & GPT
        & 2.240 & 2.440 & .585
        & 44\% & .430
        & .569 & \textbf{67\%} & \underline{.489} \\

        DialogXpert~\citep{rakib2026dialogxpert}
        & GPT
        & 2.260 & 2.564 & .603
        & \underline{45\%} & \underline{.450}
        & .548 & 63\% & .473 \\

        \midrule

        ProCoT~\citep{deng-etal-2023-prompting}
        & Qwen
        & 2.177 & 2.287 & .558
        & 36\% & .360
        & .380 & 45\% & .327 \\

        AnE~\citep{zhang2023ask}
        & Qwen
        & 1.540 & 1.717 & .407
        & 35\% & .320
        & .440 & 57\% & .365 \\

        \midrule

        Standard
        & Qwen
        & 2.083 & 2.227 & .539
        & 39\% & .390
        & .400 & 49\% & .353 \\

        SFT Init.
        & Qwen
        & 2.292 & 2.364 & .582
        & 43\% & .420
        & .531 & 55\% & .440 \\

        w/o Cog-Critic
        & Qwen
        & 2.150 & 2.244 & .549
        & 40\% & .380
        & .453 & 47\% & .389 \\
        
        w/o Cog-State
        & Qwen
        & 2.293 & 2.550 & .605
        & 45\% & .440
        & .548 & 56\% & .477 \\

        CSTPO
        & Qwen
        & 2.313 & \underline{2.679} & \underline{.624}
        & \textbf{46\%} & \textbf{.460}
        & \textbf{.581} & 57\% & \textbf{.491} \\

        \bottomrule
    \end{tabular}
    \vspace{-2mm}
    \caption{
        Detailed results on ESConv, P4G, and CB.
        \textbf{Bold} and \underline{underline} indicate the best and second-best results in each column, respectively. Appendix~\ref{app:additional-results} reports the complete Qwen3-14B results.
    }
    \label{tab:detailed-results}
    \vspace{-2mm}
\end{table*}

\textbf{Emotional Support Dialogues (ESConv).}
\textit{CSTPO} improves the SFT policy from 0.582 to 0.624 in terminal return, outperforming all Qwen-based baselines and approaching the strongest GPT-5.5 result (0.624 vs. 0.632). The improvement mainly comes from action-plan feasibility, which increases from 2.364 to 2.679, while emotion improvement remains stable (2.292 vs. 2.313). This indicates that \textit{CSTPO} learns to better advance supportive conversations toward actionable outcomes without sacrificing emotional support.

\textbf{Donation Persuasion Dialogues (P4G).}
\textit{CSTPO} achieves the highest terminal return among all compared methods on P4G, improving from 0.420 with SFT Init. to 0.460. The commitment rate also increases from 43\% to 46\%, indicating that cognitive-state-driven optimization helps the policy select more effective persuasion strategies and elicit stronger user commitment.

\textbf{Negotiation Dialogues (CraigslistBargain).}
\textit{CSTPO} achieves the largest gain on CB, improving terminal return from 0.440 with SFT Init. to 0.491 and slightly surpassing the strongest GPT-5.5 baseline (0.491 vs. 0.489). The improvement mainly comes from price utility, increasing from 0.531 to 0.581, while maintaining a comparable deal rate (55\% vs. 57\%). This result shows that \textit{CSTPO} can optimize negotiation outcomes beyond simply reaching agreements.

\subsection{Component Ablations}
\label{experiment:ablation}
\vspace{-2mm}

We remove the Critic's cognitive-state input \textbf{(\textit{w/o Cog-State})} or replace the hierarchical value functions with a shared value function \textbf{(\textit{w/o Cog-Critic})}, which removes separate credit estimation for strategy selection and utterance realization.

\begin{figure}[!htbp]
    \vspace{-2mm}
    \centering
    \includegraphics[width=0.90\linewidth]{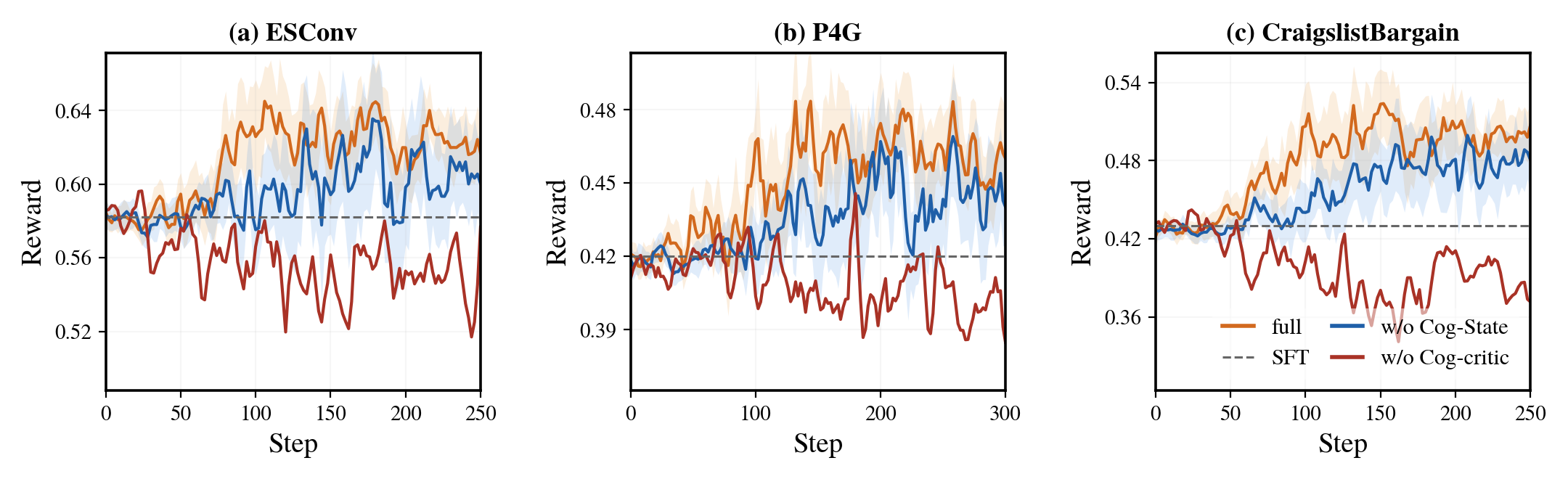}
    \vspace{-4mm}
    \caption{Qwen3-14B ablations on ESConv, P4G, and CB. Solid lines show centered three-step moving averages, shaded regions show centered interquartile bands from 50 independent realizations, and dashed lines mark SFT references. Detailed results was showed in Table~\ref{tab:detailed-results}}
    \label{fig:ablations}
    \vspace{-2mm}
\end{figure}

Removing cognitive-state inputs slows optimization and reduces final returns. The full model achieves higher tail-50-step average returns than \textit{w/o Cog-State} on ESConv (0.620 vs. 0.604), P4G (0.462 vs. 0.448), and CB (0.496 vs. 0.478), indicating that cognitive states provide additional signals for policy optimization. In contrast, \textit{w/o Cog-Critic} fails to improve over \textit{SFT init.} and remains at lower reward levels throughout training. This highlights the importance of hierarchical value estimation for fine-grained credit assignment between strategy selection and utterance realization. Appendix~\ref{app:training-dynamics} reports the Actor--Critic training traces at two model scales.

\subsection{Smaller-backbone generalization}
\label{app:generalization-8b}

\begingroup
\setlength{\columnsep}{13pt}
\setlength{\intextsep}{2pt}
\begin{wraptable}{r}{0.35\textwidth}
    \centering
    \small
    \setlength{\tabcolsep}{2pt}
    \renewcommand{\arraystretch}{1.15}
    \begin{tabular*}{\linewidth}{@{\extracolsep{\fill}}lccc@{}}
        \toprule
        \textbf{Setting} & \textbf{ESConv} & \textbf{P4G} & \textbf{CB} \\
        \midrule
        Standard & .442 & .310 & .244 \\
        SFT Init. & .494 & .350 & .321 \\
        CSTPO & \textbf{.555} & \textbf{.410} & \textbf{.435} \\
        \bottomrule
    \end{tabular*}
    \vspace{-3mm}
    \caption{Returns for the 8B model.}
    \label{tab:smaller-backbone}
\end{wraptable}

To test whether the optimization gains depend on the 14B backbone, we repeat training with an 8B Actor. Table~\ref{tab:smaller-backbone} show positive improvements over the task-specific SFT initialization on all three tasks. The largest absolute gain is on CB. The 8B endpoints remain below their 14B counterparts, so this result supports transfer across the two tested model scales rather than backbone-independent performance.

\subsection{Human Evaluation}
\label{experiment:human}
\vspace{-2mm}

We conduct human evaluation against three representative GPT-5.5 baselines from Section~\ref{experiment:main}: Standard, ProCoT and PPDPP. Five annotators perform blind pairwise comparisons, and majority voting is used for each dialogue and dimension. ESConv is evaluated on identification \textbf{(Ident.)}, comforting \textbf{(Comf.)}, and suggestion \textbf{(Sugg.)}; P4G and CB are evaluated on persuasiveness \textbf{(Pers.)}, coherence \textbf{(Coh.)}, and naturalness \textbf{(Nat.)}. We report net preference under the same setting as Section~\ref{experiment:CogSim}.
\vspace{+3mm}

\begin{table*}[ht]
    \centering
    \small
    \setlength{\tabcolsep}{1.0pt}
    \renewcommand{\arraystretch}{0.70}
    \begin{tabularx}{0.90\textwidth}{
        l
        *{9}{>{\centering\arraybackslash}X}
    }
        \toprule
        &
        \multicolumn{3}{c}{\textbf{ESConv}}
        & \multicolumn{3}{c}{\textbf{P4G}}
        & \multicolumn{3}{c}{\textbf{CB}} \\
        \cmidrule(lr){2-4}
        \cmidrule(lr){5-7}
        \cmidrule(lr){8-10}
        \textbf{Comparator}
        & Ident. & Comf. & Sugg.
        & Pers. & Coh. & Nat.
        & Pers. & Coh. & Nat. \\
        \midrule
        Standard
        & \textbf{+.23}\textsuperscript{\ensuremath{\dagger}}
        & \textbf{+.29}\textsuperscript{\ensuremath{\dagger}}
        & +.06
        & \textbf{+.13}\textsuperscript{\ensuremath{\dagger}}
        & +.05
        & \textbf{-.09}\textsuperscript{\ensuremath{\dagger}}
        & \textbf{+.15}\textsuperscript{\ensuremath{\dagger}}
        & +.07
        & \textbf{-.10}\textsuperscript{\ensuremath{\dagger}} \\
        ProCoT~\citep{deng-etal-2023-prompting}
        & \textbf{+.11}\textsuperscript{\ensuremath{\dagger}}
        & \textbf{+.13}\textsuperscript{\ensuremath{\dagger}}
        & \textbf{-.11}\textsuperscript{\ensuremath{\dagger}}
        & \textbf{+.25}\textsuperscript{\ensuremath{\dagger}}
        & \textbf{+.10}\textsuperscript{\ensuremath{\dagger}}
        & -.02
        & \textbf{+.31}\textsuperscript{\ensuremath{\dagger}}
        & \textbf{+.11}\textsuperscript{\ensuremath{\dagger}}
        & -.03 \\
        PPDPP~\citep{deng2024plug}
        & \textbf{+.15}\textsuperscript{\ensuremath{\dagger}}
        & \textbf{+.17}\textsuperscript{\ensuremath{\dagger}}
        & +.03
        & \textbf{+.21}\textsuperscript{\ensuremath{\dagger}}
        & +.07
        & -.03
        & +.07
        & +.05
        & -.04 \\
        \bottomrule
    \end{tabularx}
    \caption{Pairwise human evaluation of \textit{CSTPO} against GPT-5.5 baselines. positive values favor \textit{CSTPO}, and \ensuremath{\dagger} indicates that the 95\% dialogue-clustered bootstrap interval excludes zero.}
    \label{tab:human-preference}
\end{table*}

\vspace{+2mm}

\textit{CSTPO} obtains significant positive preferences in 13 of 27 comparisons (Table~\ref{tab:human-preference}). It consistently improves identification and comforting over all ESConv baselines and achieves significant gains in persuasiveness against all P4G baselines. On CB, \textit{CSTPO} improves persuasiveness over Standard and ProCoT. Three comparisons favor GPT-5.5 baselines, including suggestion against ProCoT on ESConv and naturalness against Standard on P4G and CB. These results suggest that \textit{CSTPO} mainly improves task-relevant dialogue abilities rather than general language quality. full details and confidence intervals are provided in Appendix~\ref{app:human-eval}.

\subsection{Strategy Adaptation after RL}
\vspace{-2mm}
We further analyze how \textit{CSTPO} changes strategy selection by comparing strategy-label distributions before and after RL. For cross-task visualization, task-specific labels are grouped into four functional categories: Advance, Soften, Probe, and Other. Detailed mappings are provided in Appendix~\ref{app:strategy}.

\begin{figure}[H]
    \centering
    \includegraphics[width=.90\linewidth]{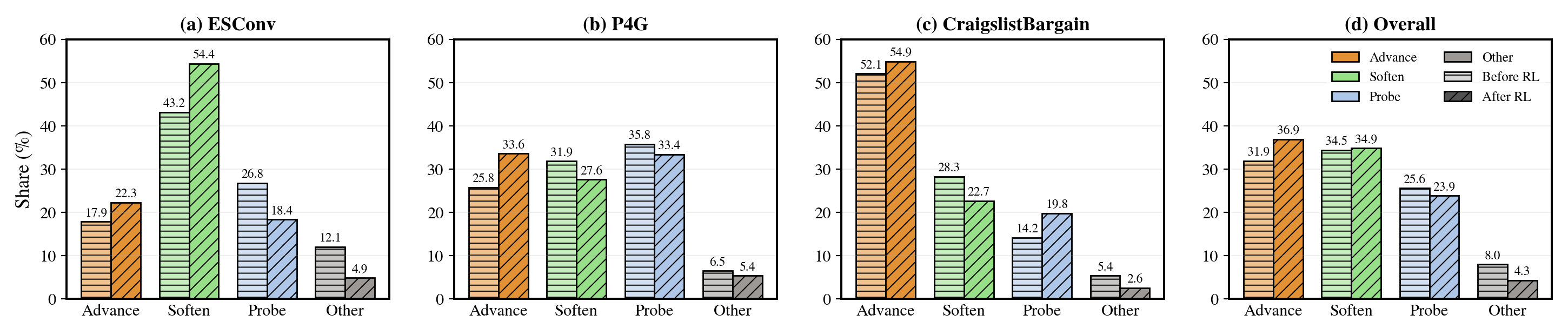}
    \vspace{-3mm}
    \caption{Strategy-label distributions before and after RL.}
    \label{fig:strategy-usage-app}
\end{figure}

\textit{CSTPO} learns task-dependent strategy redistribution (Figure~\ref{fig:strategy-usage-app}). On ESConv, \textit{CSTPO} increases Soften strategies from 43.2\% to 54.4\%, reflecting a stronger emphasis on supportive interactions. On CB, Probe strategies increase from 14.2\% to 19.8\% while preserving an Advance-dominant pattern, suggesting greater use of information-seeking behaviors during negotiation. On P4G, Advance strategies increase from 25.8\% to 33.6\%, consistent with more commitment-oriented persuasion. Meanwhile, Other strategies decrease across all tasks, indicating that \textit{CSTPO} concentrates strategy selection on task-relevant behaviors. Dialogue examples are provided in Appendix~\ref{app:cases}.

\vspace{-2mm}

%% file: sections/appendix.tex
\clearpage

\section*{Appendix Roadmap}
\label{app:roadmap}

The appendix is organized to separate the mechanism, reproducibility details,
evaluation evidence, and prompt specifications. Table~\ref{tab:appendix-roadmap}
summarizes the contents and their relation to the main paper.

\begin{table}[H]
    \centering
    \small
    \setlength{\tabcolsep}{5pt}
    \begin{tabularx}{0.96\textwidth}{c c X}
        \toprule
        \textbf{Section} & \textbf{Topic} & \textbf{Contents} \\
        \midrule
        \ref{app:cogsim} & \textit{Cog-Sim details} & State representation, cognitive routing, constrained transitions, affect updates, and recoverable checkpoints. \\
        \ref{app:cstpo} & \textit{CSTPO details} & Training algorithm, same-state continuations, Critic targets, field masks, and policy updates. \\
        \ref{app:theory} & \textit{Theoretical properties} & Training-state refinement, orthogonal hierarchical credit decomposition, and action aliasing. \\
        \ref{app:data-seeds} &\textit{ Data and seeds} & Dataset statistics, strategy spaces, seed schema, information boundaries, and seed quality control. \\
        \ref{app:implementation} & \textit{Reproducibility} & SFT and RL configurations, model backends, compute, terminal utilities, stopping rules, and statistics. \\
        \ref{app:additional-results} & \textit{Additional results} & Complete Qwen3-14B results, descriptive differences from Standard, and the fine-grained strategy mapping underlying the main-paper analysis. \\
        \ref{app:training-dynamics} & \textit{Generalization and dynamics} & Smaller-backbone results, 8B reward curves, and 14B Actor--Critic training dynamics. \\
        \ref{app:human-eval} &\textit{ Human evaluation} & Annotation protocol and complete pairwise results with confidence intervals. \\
        \ref{app:cases} & \textit{Case studies} & Selected dialogue excerpts with outcome-aware analyses and diagnostic limitations. \\
        \ref{app:prompts} & \textit{Prompts} & Actor, Cog-Sim, seed-construction, and evaluator prompts used by the proposed framework. \\
        \bottomrule
    \end{tabularx}
    \caption{Organization of the appendix.}
    \label{tab:appendix-roadmap}
\end{table}

\section{Cog-Sim Transition Details}
\label{app:cogsim}

\subsection{State representation}
\label{app:cogsim-state}

Cog-Sim represents a user by a persistent profile $P$, a dynamic cognitive
state $C_t=(B_t,D_t,I_t)$, and an affective state
$E_t=(v_t,\rho_t,c_t)$. Each belief, desire, or intention records its semantic
content, strength, centrality, activation status, and, where applicable,
direction. Strengths lie in $[0,4]$, and the simulator maintains
\begin{equation}
|B_t|\leq4,\qquad |D_t|\leq3,\qquad |I_t|\leq3.
\end{equation}
Stable processing tendencies are represented by
\begin{equation}
\Theta_u=(\eta_R,\tau_A,\tau_R),
\qquad 0\leq\tau_A<\tau_R\leq1,
\end{equation}
where $\eta_R$ controls the tendency toward systematic processing and
$\tau_A,\tau_R$ delimit acceptance and rejection regions.

\subsection{Action type, processing route, and judgment}
\label{app:cogsim-routing}

The action-type classifier maps the realized Actor utterance, rather than its
task-specific strategy label, to Influence, Elicit, or Social. Influence acts
may change BDI state; Elicit acts reveal existing state; Social acts provide
affective or relational interaction. All three modes may change affect and
produce a user response.

For an Influence act, Cog-Sim extracts relevance $Rel_t$, argument quality
$Arg_t$, peripheral-cue strength $Cue_t$, and interaction pressure $Press_t$.
The qualitative levels low, medium, and high are mapped to $.2$, $.5$, and
$.8$, respectively; the same mapping converts stance distance to $d_t$.
The central and peripheral processing scores are
\begin{equation}
S_t^{\mathrm C}=\eta_R Rel_t Arg_t,
\qquad
S_t^{\mathrm P}=(1-\eta_R)Cue_t,
\qquad
p_t^{\mathrm C}=\frac{S_t^{\mathrm C}}
{S_t^{\mathrm C}+S_t^{\mathrm P}+\varepsilon}.
\end{equation}
The central route is used when $p_t^{\mathrm C}\geq0.5$; under a weak-signal
condition, the simulator falls back to the user's stable tendency $\eta_R$.
For stance distance $d_t$, social judgment is
\begin{equation}
J_t=
\begin{cases}
\mathrm{Accept}, & d_t\leq\tau_A,\\
\mathrm{Noncommit}, & \tau_A<d_t<\tau_R,\\
\mathrm{Reject}, & d_t\geq\tau_R.
\end{cases}
\end{equation}

\subsection{Constrained cognitive transition}
\label{app:cogsim-transition}

For Influence acts, an LLM first proposes a sparse semantic change
$\widehat{\Delta C}_{t+1}$ and a response plan. A deterministic updater then
applies the route--judgment contract in Table~\ref{tab:rj-contract}, clips
strengths, enforces capacity limits, and rejects unsupported changes. Elicit
and Social preserve $C_t$. This separation lets the language model interpret
meaning while keeping the transition magnitude and direction auditable.

\begin{table}[H]
    \centering
    \scriptsize
    \setlength{\tabcolsep}{4pt}
    \begin{tabularx}{0.97\textwidth}{llXccc}
        \toprule
        \textbf{Route} & \textbf{Judgment} & \textbf{Permitted change} & $|\Delta B|$ & $|\Delta D|$ & $|\Delta I|$ \\
        \midrule
        Central & Accept & Substantive evidence may change directly relevant beliefs; desires and intentions may move when supported. & 1.0 & .5 & .8 \\
        Central & Noncommit & Relevant beliefs may loosen without reversal; desires remain stable and no strong commitment is created. & .4 & .2 & .3 \\
        Central & Reject & The rejected claim cannot be reinforced; counter-beliefs may emerge and a target intention may weaken. & .4 & .2 & .5 \\
        Peripheral & Accept & Cue-related trust or norm beliefs may strengthen; core issue beliefs and desires remain nearly fixed. & .3 & .1 & .6 \\
        Peripheral & Noncommit & Only small cue-driven belief or intention movement is permitted. & .2 & 0 & .2 \\
        Peripheral & Reject & Cue-related trust may weaken; core beliefs and desires remain stable and a target intention may decrease. & .2 & 0 & .5 \\
        \bottomrule
    \end{tabularx}
    \caption{Summary of the route--judgment transition contracts. Bounds denote the maximum absolute strength change per affected node.}
    \label{tab:rj-contract}
\end{table}

The affect engine evaluates goal congruence $GC_t$, coping potential $CP_t$,
and future expectancy $FE_t$ after the constrained cognitive update:
\begin{equation}
\widehat v_{t+1}=\frac{GC_t+CP_t+FE_t}{3},
\qquad
\widehat\rho_{t+1}=0.6|\Delta C_t|+0.25|\Delta GC_t|+0.5Press_t.
\end{equation}
The state is smoothed as
\begin{equation}
v_{t+1}=(1-\lambda_q)v_t+\lambda_q\widehat v_{t+1},
\qquad
\rho_{t+1}=(1-\lambda_q)\rho_t+\lambda_q\widehat\rho_{t+1},
\end{equation}
where Influence and Elicit use $\lambda_q=.4$ and Social uses
$\lambda_q=.25$. The final response is generated from the updated state and a
mode-specific response plan without exposing internal state terminology.

\subsection{Recoverable checkpoints}
\label{app:cogsim-checkpoints}

A pre-action checkpoint contains the complete user state, dialogue history,
profile, cognitive parameters, previous goal congruence, task events,
termination status, remaining horizon, and random stream. Every continuation
at a branch parent restores this checkpoint before sampling an alternative
action, so candidate continuations share the same observed prefix and latent
user state.

\section{CSTPO Training Details}
\label{app:cstpo}

\begin{algorithm}[H]
\caption{Cognitive-State Transition--Driven Policy Optimization (CSTPO)}
\label{alg:cstpo}
\begin{algorithmic}[1]
\REQUIRE SFT Actor $\pi_{\mathrm{SFT}}$, Critics $V_\phi,U_\phi$,
Cog-Sim, seed dataset $\mathcal D$, branch rule $\mathsf{Select}$,
strategy count $m$, utterance count $\ell$
\ENSURE Optimized policy $\pi_\theta$
\STATE Initialize $\pi_\theta\leftarrow\pi_{\mathrm{ref}}$ and freeze $\pi_{\mathrm{ref}}$
\STATE Initialize $V_\phi,U_\phi$
\WHILE{not converged}
    \STATE Freeze $\pi_{\mathrm{old}}\leftarrow\pi_\theta$ and
    $(V_{\mathrm{old}},U_{\mathrm{old}})\leftarrow(V_\phi,U_\phi)$
    \STATE Sample a batch of seeds and set $\mathcal B\leftarrow\emptyset$
    \FOR{each seed}
        \WHILE{the main trajectory is non-terminal}
            \STATE Save $(x_t,o_t)$; if $\mathsf{Select}(x_t)=1$, add its
            checkpoint to $\mathcal B$ before sampling the current action
            \STATE Sample $(k_t,u_t)\sim\pi_{\mathrm{old}}(\cdot\mid o_t)$,
            store old log probabilities and field masks, and step Cog-Sim
        \ENDWHILE
        \STATE Evaluate and store the main terminal return $G(\tau)$
    \ENDFOR
    \FOR{each branch parent $b\in\mathcal B$}
    \STATE Index the stored main continuation as
    $(k_{b,1},u_{b,1,1},G_{b,1,1})$
    \STATE Sample $m-1$ additional distinct strategies
    $\{k_{b,i}\}_{i=2}^{m}$ under $\pi_{\mathrm{old}}(\cdot\mid o_b)$
    \FOR{$i=1,\ldots,m$}
        \FOR{$r=1,\ldots,\ell$}
            \IF{$(i,r)\neq(1,1)$}
                \STATE Restore $x_b$, condition on $k_{b,i}$, sample
                $u_{b,i,r}$ under $\pi_{\mathrm{old}}$, and continue to termination
                \STATE Store the terminal return $G_{b,i,r}$
            \ENDIF
        \ENDFOR
        \STATE $\overline G_{b,i}\leftarrow
        \ell^{-1}\sum_{r=1}^{\ell}G_{b,i,r}$
    \ENDFOR
    \ENDFOR
    \FOR{each main-path turn $t$}
        \IF{$t$ is an expanded parent $b$}
            \STATE $\overline G_{t,1}\leftarrow\overline G_{b,1}$
        \ELSE
            \STATE $\overline G_{t,1}\leftarrow G(\tau)$
        \ENDIF
        \STATE $\widehat A_t^{\,k}\leftarrow
        \overline G_{t,1}-V_{\mathrm{old}}(x_t)$
        \STATE $\widehat A_t^{\,u}\leftarrow
        G(\tau)-U_{\mathrm{old}}(x_t,k_t)$
    \ENDFOR
    \STATE Fit $V_\phi,U_\phi$ using Eqs.~\ref{eq:critic-main-app}--\ref{eq:critic-full-app}
    \STATE Update $\pi_\theta$ with field-aware PPO on main-path action
    tokens only, using ratios against $\pi_{\mathrm{old}}$
\ENDWHILE
\RETURN $\pi_\theta$
\end{algorithmic}
\end{algorithm}

\subsection{Critic objective}
\label{app:cstpo-critic}

Let $\mathcal D_{\mathrm{main}}$ be the set of main-path pre-action nodes and
$\mathcal B$ the set of expanded parents. For parent $b$, let
$\mathcal E_{b,k}$ contain the terminal returns of additional continuations
that sampled strategy $k$, and let
$\mathcal K_b^+=\{k:|\mathcal E_{b,k}|>0\}$. Main-path nodes supervise both
value heads:
\begin{equation}
\mathcal L_{\mathrm{main}}(\phi)=
\frac{1}{|\mathcal D_{\mathrm{main}}|}
\sum_{(x,k,G)\in\mathcal D_{\mathrm{main}}}
\left[(V_\phi(x)-G)^2+(U_\phi(x,k)-G)^2\right].
\label{eq:critic-main-app}
\end{equation}
Additional continuations supervise only the sampled strategy-conditioned value
at their branch parent:
\begin{equation}
\mathcal L_{\mathrm{branch}}(\phi)=
\frac{1}{|\mathcal B|}\sum_{b\in\mathcal B}
\frac{1}{|\mathcal K_b^+|}\sum_{k\in\mathcal K_b^+}
\frac{1}{|\mathcal E_{b,k}|}\sum_{G\in\mathcal E_{b,k}}
(U_\phi(x_b,k)-G)^2.
\end{equation}
The complete objective is
\begin{equation}
\mathcal L_{\mathrm{Critic}}(\phi)=
\mathcal L_{\mathrm{main}}(\phi)
+\lambda_{\mathrm{br}}\mathcal L_{\mathrm{branch}}(\phi),
\qquad \lambda_{\mathrm{br}}=.25.
\label{eq:critic-full-app}
\end{equation}
The main continuation is counted only in $\mathcal L_{\mathrm{main}}$, and an
unsampled strategy never receives an artificial zero target. The frozen old
Critic forms advantages before the current batch is fitted.

\subsection{Sampling and action masks}
\label{app:cstpo-sampling}

Each RL step samples 10 seeds and generates one main trajectory per seed. We select two non-terminal pre-action checkpoints per main trajectory for P4G and CB and four for ESConv. At each selected checkpoint, we use $m=2$ distinct strategies and sample $\ell=2$ utterance realizations for each strategy under the frozen old policy $\pi_{\mathrm{old}}$. The stored main continuation supplies one of the four strategy--utterance paths, so only three additional tails are generated. This yields seven rollouts per seed for P4G and CB and thirteen for ESConv. All four continuations share the same pre-action state, Cog-Sim version, remaining horizon, and terminal evaluator. Parent selection is committed before the current action, next user response, or terminal return is observed.

Only main-path strategy and utterance tokens enter the Actor loss. System prompts, user messages, forced formatting tokens, and all additional continuation tokens have zero Actor mask. Old and new strategy probabilities are evaluated on the same trie-constrained label support, while utterance probabilities use the regular vocabulary.

\section{Theoretical Properties of CSTPO}
\label{app:theory}

The following results formalize the population quantities that motivate the
training-side state and the hierarchical Critic. Let $G\in L^2$ be the
terminal return under a fixed policy. Let $O$, $X$, $(X,K)$, and $(X,K,U)$
denote successively richer information sets, where $O=\mathcal O(X)$ is the
Actor-visible observation, $X$ is the training-side state, and $K,U$ are the
strategy and utterance. Thus their generated sigma-algebras are nested. Define
\begin{equation}
\begin{aligned}
H(O)&=\mathbb E[G\mid O], &
V(X)&=\mathbb E[G\mid X],\\
U(X,K)&=\mathbb E[G\mid X,K], &
Q(X,K,U)&=\mathbb E[G\mid X,K,U].
\end{aligned}
\label{eq:theory-values-app}
\end{equation}

\subsection{Training-state refinement}
\label{app:theory-state}

\paragraph{Proposition 1.}
The ideal value based on the training-side state satisfies
\begin{equation}
\mathbb E[(G-H(O))^2]
=
\mathbb E[(G-V(X))^2]
+
\mathbb E[(V(X)-H(O))^2].
\label{eq:state-refinement-app}
\end{equation}
Consequently, $\mathbb E[(G-V(X))^2]\leq
\mathbb E[(G-H(O))^2]$.

\paragraph{Proof.}
Write $G-H(O)=(G-V(X))+(V(X)-H(O))$. Because
$\mathbb E[G-V(X)\mid X]=0$ and $V(X)-H(O)$ is measurable with respect to
$X$, the cross term has zero expectation:
\begin{equation}
\mathbb E[(G-V(X))(V(X)-H(O))]
=
\mathbb E[(V(X)-H(O))\mathbb E[G-V(X)\mid X]]=0.
\end{equation}
Expanding the square gives Eq.~\ref{eq:state-refinement-app}. \hfill$\square$

The nonnegative gap
$\Delta_{\mathrm{state}}=\mathbb E[(V(X)-H(O))^2]$ measures the additional
return-predictive information in the full training state. Because $X$ also
contains task and horizon variables, this identity does not attribute the gap
exclusively to cognition; the ablation study isolates the empirical role of
the cognitive-affective fields.

\subsection{Hierarchical orthogonal decomposition}
\label{app:theory-hierarchy}

\paragraph{Proposition 2.}
The joint-action advantage decomposes as
\begin{equation}
Q(X,K,U)-V(X)
=
\underbrace{U(X,K)-V(X)}_{A^K(X,K)}
+
\underbrace{Q(X,K,U)-U(X,K)}_{A^U(X,K,U)},
\label{eq:orthogonal-decomposition-app}
\end{equation}
where the two components are orthogonal in expectation:
\begin{equation}
\mathbb E[A^K A^U]=0,\qquad
\mathbb E[(Q-V)^2]=\mathbb E[(A^K)^2]+\mathbb E[(A^U)^2].
\label{eq:orthogonal-components-app}
\end{equation}

\paragraph{Proof.}
Equation~\ref{eq:orthogonal-decomposition-app} follows by adding and
subtracting $U(X,K)$. By the tower property,
$U(X,K)=\mathbb E[Q(X,K,U)\mid X,K]$, and hence
$\mathbb E[A^U\mid X,K]=0$. Since $A^K$ is measurable with respect to
$(X,K)$,
\begin{equation}
\mathbb E[A^K A^U]
=\mathbb E[A^K\mathbb E[A^U\mid X,K]]=0.
\end{equation}
Expanding the squared sum yields Eq.~\ref{eq:orthogonal-components-app}.
\hfill$\square$

\subsection{Action aliasing and the unified decomposition}
\label{app:theory-aliasing}

If an action is represented only by its strategy label, the omitted
within-strategy value variation is
\begin{equation}
\Delta_{\mathrm{alias}}
=\mathbb E[\operatorname{Var}(Q(X,K,U)\mid X,K)]
=\mathbb E[(Q(X,K,U)-U(X,K))^2].
\label{eq:action-aliasing-app}
\end{equation}
A positive $\Delta_{\mathrm{alias}}$ means that utterances sharing a strategy
label can have different long-term values, motivating an explicit utterance
component rather than a label-only action space.

Together, the nested conditional expectations give
\begin{equation}
\begin{aligned}
G-H(O)
={}&[V(X)-H(O)]+[U(X,K)-V(X)]\\
&+[Q(X,K,U)-U(X,K)]+[G-Q(X,K,U)].
\end{aligned}
\label{eq:unified-decomposition-app}
\end{equation}

The four increments are pairwise orthogonal under the fixed-policy trajectory
distribution. Their squared expectations therefore add to
$\mathbb E[(G-H(O))^2]$. This population identity motivates progressively
conditioning the value estimate on training state, strategy, and utterance.
It does not assert that learned Critics equal the conditional expectations,
that the finite-sample advantages sum exactly to a joint advantage, or that
PPO converges to an optimal policy.

\section{Data and Interaction Seeds}
\label{app:data-seeds}

\subsection{Datasets and action spaces}
\label{app:data-action-spaces}

\begin{table}[H]
    \centering
    \small
    \setlength{\tabcolsep}{5pt}
    \begin{tabularx}{0.96\textwidth}{lcccX}
        \toprule
        \textbf{Task} & \textbf{Train/Dev/Test} & \textbf{Actor/User} & \textbf{Strategies} & \textbf{Task objective} \\
        \midrule
        ESConv & 1196/150/150 & supporter/seeker & 8 & Improve the user's emotion or hope and support a feasible action plan. \\
        P4G & 813/102/102 & persuader/persuadee & 13 & Obtain a clear, voluntary donation commitment that is not withdrawn. \\
        CB & 5247/597/838 & buyer/seller & 6 & Reach a valid agreement while optimizing the buyer-side price utility. \\
        \bottomrule
    \end{tabularx}
    \caption{Datasets, roles, and task-specific strategy spaces. Policies are trained separately and do not share strategy labels or reward scales.}
    \label{tab:datasets-app}
\end{table}

For each task, we sample 200/50/100 high-quality dialogue seeds for policy
training, development, and test-time evaluation. The ESConv labels are
Question, Restatement or Paraphrasing, Reflection of
Feelings, Self-disclosure, Affirmation and Reassurance, Providing Suggestions,
Information, and Others. P4G uses 13 normalized labels covering donation
requests, logical, emotional, and credibility appeals, donation information,
personal stories, self-modeling, foot-in-the-door, praise, inquiry,
inquiry response, social interaction, and other. CB uses propose-price,
inquire, inform, stance, social, and other; structured offer, acceptance, and
termination events are stored separately from strategy labels.

\subsection{Seed schema and information boundaries}
\label{app:seed-schema}

Each seed records provenance and group identifiers, both dialogue roles,
separate Actor and user task views, a dialogue prefix, the user profile,
cognitive parameters $\Theta_u$, initial BDI and affective states, the random
stream, turn budget, and component versions. ESConv additionally stores the
problem situation and initial affective evidence; P4G stores task-relevant
questionnaire attributes; CB stores the item knowledge base and each party's
private target price.

The Actor receives only its role-visible task view and dialogue history.
Cog-Sim receives the simulated user's private task view and cognitive state.
The Critic may access the training-side pre-action context $x_t$, but neither
the Actor nor terminal evaluator receives latent user state. The evaluator
receives only the completed dialogue and whitelisted case facts required by
the task rubric.

\subsection{State annotation and quality control}
\label{app:seed-quality}

Offline annotation converts source-dialogue evidence into a draft initial
profile, BDI state, and affective state. It distinguishes states already
observed at the rollout cutoff, stable dispositions revealed elsewhere in the
source dialogue, and reactive changes caused by later conversational actions;
reactive changes are excluded. The rollout seed contains the audited initial
state rather than future source utterances or source outcomes.

An automatic first pass checks semantic type, first-person formulation,
evidence provenance, role attribution, cutoff legality, and task-specific
constraints. Human review then verifies source grounding, absence of outcome
leakage, role visibility, Persona/BDI separation, structural validity,
coverage, parameter validity, Persona utility, and whether the state changes
only when supported by evidence. Seeds that fail review are corrected and
re-reviewed or excluded.

\section{Implementation and Evaluation Details}
\label{app:implementation}

\subsection{SFT and reinforcement learning}
\label{app:training-setup}

We train a separate Qwen3-14B Actor for each task. SFT targets are serialized as a strategy label, a newline, and the utterance. Human dialogues and frozen teacher trajectories are mixed at a 1:1 ratio. SFT uses LoRA with rank 64, $\alpha=128$, dropout .05, maximum sequence length 4096, learning rate $2\times10^{-5}$, and four epochs.

RL initializes from the task-specific SFT checkpoint and uses a separate Critic; no KL penalty to an SFT reference policy is applied. At each PPO iteration, $\pi_{\mathrm{old}}$ is a frozen snapshot of the current Actor and serves as both the rollout policy and likelihood-ratio denominator. The Actor and Critic learning rates are $1\times10^{-5}$ and $1\times10^{-4}$, respectively, and both use \textit{LoRA} rank 64. We use a train batch of 10 seeds, a PPO mini-batch of 10, clipping coefficient .2, and $\gamma=1$. ESConv and CB are trained for 250 optimization steps, while P4G is trained for 300 steps. Rollouts use temperature 1, top-$p$ 1, trie-constrained strategy generation, at most 128 generated tokens per Actor utterance, and at most 30 Actor--User turns. Training uses \texttt{verl} and \texttt{vLLM}, bf16 arithmetic, and Qwen3 thinking disabled. All training runs use two NVIDIA A100 80GB GPUs.

Cog-Sim, the baseline simulators used for simulator validation, and the terminal evaluator use the frozen \texttt{DeepSeek-V4-Pro-0813} backend. Each terminal judgment is sampled independently three times. The evaluator never observes Cog-Sim hidden states or Critic outputs.

\subsection{Termination and terminal utilities}
\label{app:terminal-utilities}

Dialogues may end naturally. Otherwise, redundancy is checked every two turns
from turn 13, and all conversations have a hard limit of 30 turns. Numeric
Judge fields are averaged across three judgments; Boolean fields use majority
vote and nullable extracted fields use the non-null mode.

For ESConv, the evaluator assigns emotion/hope improvement $E\in[0,4]$ and
action-plan feasibility $A\in[0,4]$:
\begin{equation}
G_{\mathrm{ESConv}}=\frac{E+A}{8}.
\end{equation}
For P4G, $G_{\mathrm{P4G}}=1$ if the user makes a clear, voluntary,
non-purely-conditional commitment that is not withdrawn; otherwise it is zero.
For CB, with seller target $p_s$, buyer target $p_b$, and final price $p$,
\begin{equation}
SL=\begin{cases}
\dfrac{p-p_s}{p_b-p_s}, & \text{valid deal with an extractable price},\\[4pt]
0, & \text{otherwise},
\end{cases}
\qquad
G_{\mathrm{CB}}=\operatorname{clip}(SL,0,1).
\end{equation}
All paired comparisons use 2,000 case-clustered bootstrap samples and 95\%
confidence intervals.

\subsection{Cog-Sim validation protocol}
\label{app:cogsim-validation}

We evaluate Cog-Sim on 100 frozen seeds per task. From the same pre-action
checkpoint, each simulator receives four ordered input levels and two
paraphrases per level. ESConv and P4G progress from absent or weak evidence to
relevant evidence and an actionable proposal. In CB, L0 asks without an offer,
L1 offers 95\% of the current asking price, L2 offers 98--99\%, and L3 accepts
the full asking price with an immediate-payment commitment. A frozen LLM Judge
scores response change on a 0--3 scale. Randomized pairwise evaluation reports
naturalness, consistency, and dose appropriateness as net preference
$(W-L)/N$, with ties included in $N$.

\section{Additional Experimental Results}
\label{app:additional-results}

\subsection{Complete Qwen3-14B results}
\label{app:qwen-results}

Table~\ref{tab:qwen-full-app} reports all evaluated Qwen3-14B policies. The
main paper presents the principal cross-backbone comparison; this table adds
the remaining Qwen3-14B results without introducing additional metrics.

\begin{table}[H]
    \centering
    \small
    \setlength{\tabcolsep}{4pt}
    \renewcommand{\arraystretch}{1.15}
    \hypersetup{pdfborder={0 0 0}}
    \begin{tabular*}{\linewidth}{@{\extracolsep{\fill}}lrrr@{\hspace{10pt}}rr@{\hspace{10pt}}rrr@{}}
        \toprule
        & \multicolumn{3}{c}{\textbf{ESConv}} & \multicolumn{2}{c}{\textbf{P4G}} & \multicolumn{3}{c}{\textbf{CB}} \\
        \cmidrule(lr){2-4}\cmidrule(lr){5-6}\cmidrule(lr){7-9}
        \textbf{Method} & $E$ & $A$ & $G$ & Comm. & $G$ & SL & Deal & $G$ \\
        \midrule
        Standard
        & 2.083 & 2.227 & .539
        & 39\% & .390
        & .400 & 49\% & .353 \\

        Proactive~\citep{deng-etal-2023-prompting}
        & 2.127 & 1.860 & .498
        & 31\% & .310
        & .495 & 54\% & .412 \\

        ProCoT~\citep{deng-etal-2023-prompting}
        & 2.177 & 2.287 & .558
        & 36\% & .360
        & .386 & 45\% & .327 \\

        AnE~\citep{zhang2023ask}
        & 1.540 & 1.717 & .407
        & 35\% & .320
        & .443 & \underline{57\%} & .365 \\

        MI-Prompt~\citep{chen-etal-2023-controllable}
        & 1.833 & 1.490 & .415
        & 42\% & \underline{.420}
        & .442 & \textbf{61\%} & .370 \\

        PPDPP~\citep{deng2024plug}
        & 2.110 & 2.210 & .540
        & 36\% & .350
        & .460 & 52\% & .375 \\

        DialogXpert~\citep{rakib2026dialogxpert}
        & 2.150 & 2.270 & .552
        & 37\% & .370
        & .482 & 54\% & .389 \\

        \midrule
        SFT Init.
        & \underline{2.292}
        & \underline{2.364}
        & \underline{.582}
        & \underline{43\%}
        & \underline{.420}
        & \underline{.531}
        & 55\%
        & \underline{.440} \\

        \textit{CSTPO}
        & \textbf{2.313}
        & \textbf{2.679}
        & \textbf{.624}
        & \textbf{46\%}
        & \textbf{.460}
        & \textbf{.581}
        & \underline{57\%}
        & \textbf{.491} \\
        \bottomrule
    \end{tabular*}
    \caption{Complete Qwen3-14B evaluation. \textbf{Bold} and \underline{underlined} values denote the best and second-best results in each column, respectively.}
    \label{tab:qwen-full-app}
\end{table}

\subsection{Differences from the Qwen3-14B Standard policy}
\label{app:mean-differences}

\begingroup
\setlength{\columnsep}{12pt}
\setlength{\intextsep}{0pt}
\begin{wraptable}{r}{0.44\textwidth}
    \centering
    \small
    \setlength{\tabcolsep}{4pt}
    \renewcommand{\arraystretch}{1.15}
    \begin{tabular*}{\linewidth}{@{\extracolsep{\fill}}lrrr@{}}
        \toprule
        \textbf{Method} & \textbf{ESConv} & \textbf{P4G} & \textbf{CB} \\
        \midrule
        SFT Init. & $+.043$ & $+.030$ & $+.087$ \\
        CSTPO & $+.085$ & $+.070$ & $+.138$ \\
        \bottomrule
    \end{tabular*}
    \vspace{-2mm}
    \caption{\raggedright Return over \mbox{Qwen3-14B} Standard.}
    \label{tab:paired-standard-app}
\end{wraptable}

Table~\ref{tab:paired-standard-app} reports descriptive differences in terminal
return computed directly from the means in Table~\ref{tab:qwen-full-app}.
CSTPO improves over the Qwen3-14B Standard policy on all three tasks, with the
largest absolute difference on CB. These mean differences are descriptive and
are not used as separate significance tests.
\par
\ifnum\value{WF@wrappedlines}>1
    \vspace{\dimexpr\value{WF@wrappedlines}\baselineskip-\baselineskip\relax}
\fi
\WFclear
\endgroup
\vspace{-0.5\baselineskip}

\subsection{Strategy mapping and fine-grained shifts}
\label{app:strategy}

The main paper reports strategy redistribution at the level of four functional
groups. Table~\ref{tab:strategy-mapping-app} gives the mapping from all 27
task-specific labels to these groups. Tables~\ref{tab:strategy-def-esconv-app}--
\ref{tab:strategy-def-cb-app} then define the individual labels. The grouping
is used only for cross-task analysis and does not alter the policy's action
space.

\begin{table}[t]
    \centering
    \small
    \setlength{\tabcolsep}{6pt}
    \renewcommand{\arraystretch}{1.12}
    \begin{tabularx}{0.9\linewidth}{@{}p{.13\linewidth}>{\raggedright\arraybackslash}X@{}}
        \toprule
        \textbf{Task} & \textbf{Strategy labels} \\
        \midrule
        \multicolumn{2}{@{}l@{}}{\textbf{Advance} --- Directly advance the task objective} \\[2pt]
        ESConv & \textit{Providing Suggestions} \\
        P4G & \textit{Donate Request}; \textit{Logical Appeal}; \textit{Credibility Appeal}; \textit{Foot in the Door} \\
        CB & \textit{Propose Price}; \textit{Stance} \\
        \addlinespace[6pt]
        
        \multicolumn{2}{@{}l@{}}{\textbf{Soften} --- Build rapport and reduce interactional tension} \\[2pt]
        ESConv & \textit{Affirmation and Reassurance}; \textit{Reflection of Feelings}; \textit{Self-disclosure} \\
        P4G & \textit{Praise User}; \textit{Social}; \textit{Emotion Appeal}; \textit{Self Modeling}; \textit{Personal Story} \\
        CB & \textit{Social} \\
        \addlinespace[6pt]
        
        \multicolumn{2}{@{}l@{}}{\textbf{Probe} --- Obtain or exchange information before advancing} \\[2pt]
        ESConv & \textit{Question}; \textit{Information}; \textit{Restatement or Paraphrasing} \\
        P4G & \textit{Donation Information}; \textit{Inquiry}; \textit{Inquiry Response} \\
        CB & \textit{Inquire}; \textit{Inform} \\
        \addlinespace[6pt]
        
        \multicolumn{2}{@{}l@{}}{\textbf{Other} --- Residual or generic acts} \\[2pt]
        ESConv & \textit{Others} \\
        P4G & \textit{Other} \\
        CB & \textit{Other} \\
        \bottomrule
    \end{tabularx}
    \vspace{-2mm}
    \caption{Post-hoc mapping from task-specific strategy labels to four cross-task groups.}
    \label{tab:strategy-mapping-app}
\end{table}

\begin{table}[t]
    \centering
    \small
    \setlength{\tabcolsep}{5pt}
    \renewcommand{\arraystretch}{1.12}
    \begin{tabularx}{0.97\textwidth}{
        >{\hsize=0.6\hsize\raggedright\arraybackslash}X
        >{\hsize=1.4\hsize\raggedright\arraybackslash}X
    }
        \toprule
        \textbf{Strategy} & \textbf{Definition} \\
        \midrule
        \textit{Question} &
        Ask about the user's situation, feelings, needs, or preferences to elicit relevant information. \\

        \textit{Restatement or Paraphrasing} &
        Restate the user's expressed content in different words to confirm or clarify understanding. \\

        \textit{Reflection of Feelings} &
        Identify and reflect the emotion conveyed by the user. \\

        \textit{Self-disclosure} &
        Share a relevant personal experience or feeling to build rapport. \\

        \textit{Affirmation and Reassurance} &
        Validate the user's feelings or strengths and provide encouragement or reassurance. \\

        \textit{Providing Suggestions} &
        Recommend concrete coping actions or feasible next steps. \\

        \textit{Information} &
        Provide factual or explanatory information relevant to the user's problem. \\

        \textit{Others} &
        Use a supportive act that is not captured by the preceding categories. \\
        \bottomrule
    \end{tabularx}
    \vspace{-2mm}
    \caption{ESConv strategy definitions and post-hoc functional groups.}
    \label{tab:strategy-def-esconv-app}
\end{table}

\begin{table}[H]
    \centering
    \small
    \setlength{\tabcolsep}{5pt}
    \renewcommand{\arraystretch}{1.12}
    \begin{tabularx}{0.97\textwidth}{
        >{\hsize=0.4\hsize\raggedright\arraybackslash}X
        >{\hsize=1.6\hsize\raggedright\arraybackslash}X
    }
        \toprule
        \textbf{Strategy} & \textbf{Definition} \\
        \midrule
        \textit{Donate Request} &
        Propose, request, confirm, or negotiate a donation, including its amount or reasons for declining. \\
        \textit{Logical Appeal} &
        Use reasons or evidence to explain why donating is useful or impactful. \\
        \textit{Credibility Appeal} &
        Establish the charity's trustworthiness, reputation, or accountability. \\
        \textit{Foot in the Door} &
        Seek a small initial commitment that can facilitate a subsequent donation. \\
        \textit{Praise User} &
        Compliment the user or affirm the user's prosocial qualities. \\
        \textit{Social} &
        Perform conversational functions such as greeting, acknowledgment, thanks, or closing. \\
        \textit{Emotion Appeal} &
        Evoke empathy or compassion by emphasizing emotional consequences. \\
        \textit{Self Modeling} &
        Present the persuader's own donation or intended action as an example. \\
        \textit{Personal Story} &
        Use a personal or beneficiary-centered narrative to motivate donation. \\
        \textit{Donation Information} &
        Provide factual information about the charity, campaign, donation process, or use of funds. \\
        \textit{Inquiry} &
        Ask a task-, person-, or source-related question to understand the user. \\
        \textit{Inquiry Response} &
        Answer or otherwise respond to an inquiry raised during the donation discussion. \\
        \textit{Other} &
        Use an off-task or otherwise unclassified dialogue act. \\
        \bottomrule
    \end{tabularx}
    \caption{P4G strategy definitions and post-hoc functional groups.}
    \label{tab:strategy-def-p4g-app}
\end{table}

\begin{table}[H]
    \centering
    \small
    \setlength{\tabcolsep}{5pt}
    \renewcommand{\arraystretch}{1.12}
    \begin{tabularx}{0.97\textwidth}{
        >{\hsize=0.4\hsize\raggedright\arraybackslash}X
        >{\hsize=1.6\hsize\raggedright\arraybackslash}X
    }
        \toprule
        \textbf{Strategy} & \textbf{Definition} \\
        \midrule
        \textit{Propose Price} &
        Make an initial, counter, or deliberately underspecified price proposal. \\

        \textit{Stance} &
        Express agreement, disagreement, or insistence toward a negotiating position. \\

        \textit{Social} &
        Greet the counterpart or perform another rapport-oriented conversational act. \\

        \textit{Inquire} &
        Request information about the item, transaction, or counterpart's preferences. \\

        \textit{Inform} &
        Provide relevant item, transaction, preference, or constraint information without proposing a price. \\

        \textit{Other} &
        Use a negotiation act that is not captured by the preceding categories. \\
        \bottomrule
    \end{tabularx}
    \caption{CraigslistBargain strategy definitions and post-hoc functional groups. Structured offer, acceptance, rejection, and termination events are recorded separately.}
    \label{tab:strategy-def-cb-app}
\end{table}

\begin{figure}[H]
    \centering
    \includegraphics[width=.96\linewidth]{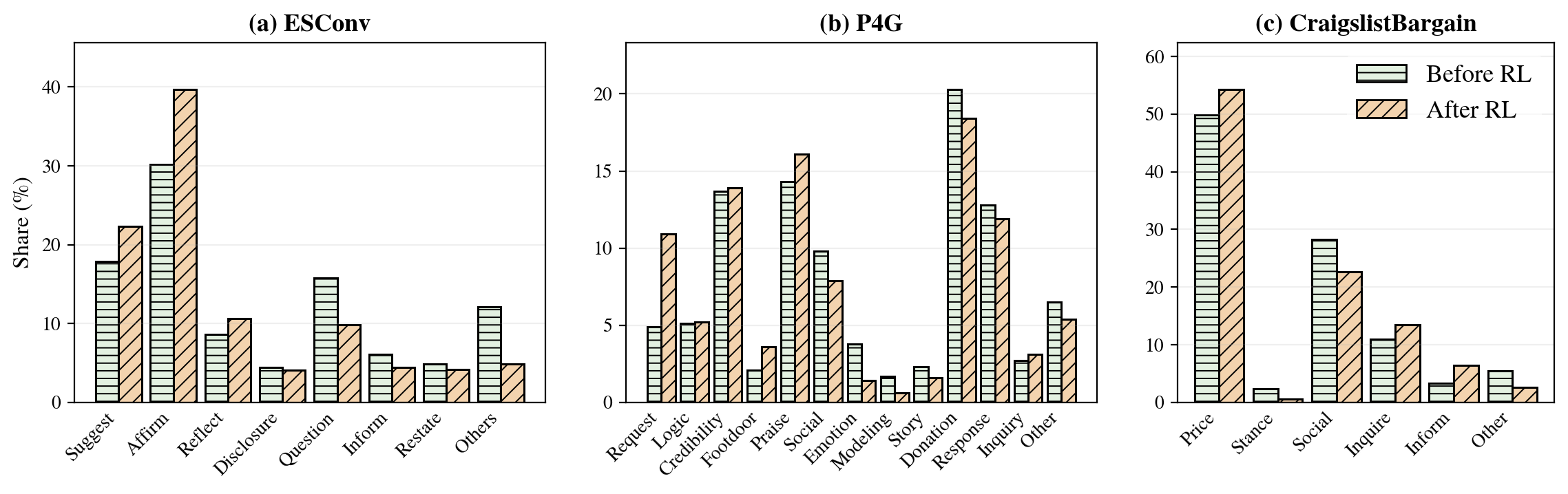}
    \vspace{-2mm}
    \caption{Fine-grained task-specific strategy distributions before and after RL.}
    \label{fig:strategy-fine-app}
\end{figure}

Figure~\ref{fig:strategy-fine-app} reports the underlying labels. On ESConv, \textit{Affirmation and Reassurance} increases by 9.5 points, while \textit{Question} and \textit{Others} decrease by 6.0 and 7.2 points. On CB, \textit{Propose Price} and \textit{Inform} increase by 4.5 and 3.1 points, whereas \textit{social} decreases by 5.6 points. On P4G, \textit{Donate Request} has the largest increase at 6.0 points. These shifts describe how the policy reallocates strategy use; they do not identify the causal contribution of a label or measure the quality of its utterance realization.

\section{Generalization and Training Dynamics}
\label{app:training-dynamics}

\paragraph{Curve processing.}
For a sequence $y_0,\ldots,y_T$, we use the centered moving average

\begin{equation}
\widetilde y_t^{(k)}=
\frac{1}{|\mathcal W_t^{(k)}|}
\sum_{s\in\mathcal W_t^{(k)}}y_s,
\qquad
\mathcal W_t^{(k)}=\{s:\max(0,t-k)\le s\le\min(T,t+k)\}.
\end{equation}

Reward curves use $k=1$ and loss curves use $k=2$, corresponding to windows of at most three and five steps, respectively; boundary windows are truncated to the available steps. Platform values are the mean of the final 50 points of the smoothed reward trajectory. For each step $t$, we draw 50 independent realizations from $\mathcal N(\mu_t,\sigma_t^2)$ and compute their 25th and 75th percentiles, $Q_{.25,t}$ and $Q_{.75,t}$. The shaded band is centered on $\widetilde y_t^{(1)}$ with half-width $(Q_{.75,t}-Q_{.25,t})/2$.s

\subsection{Smaller-backbone generalization}
\label{app:generalization-8b}

\begin{figure}[H]
    \centering
    \includegraphics[width=.85\linewidth]{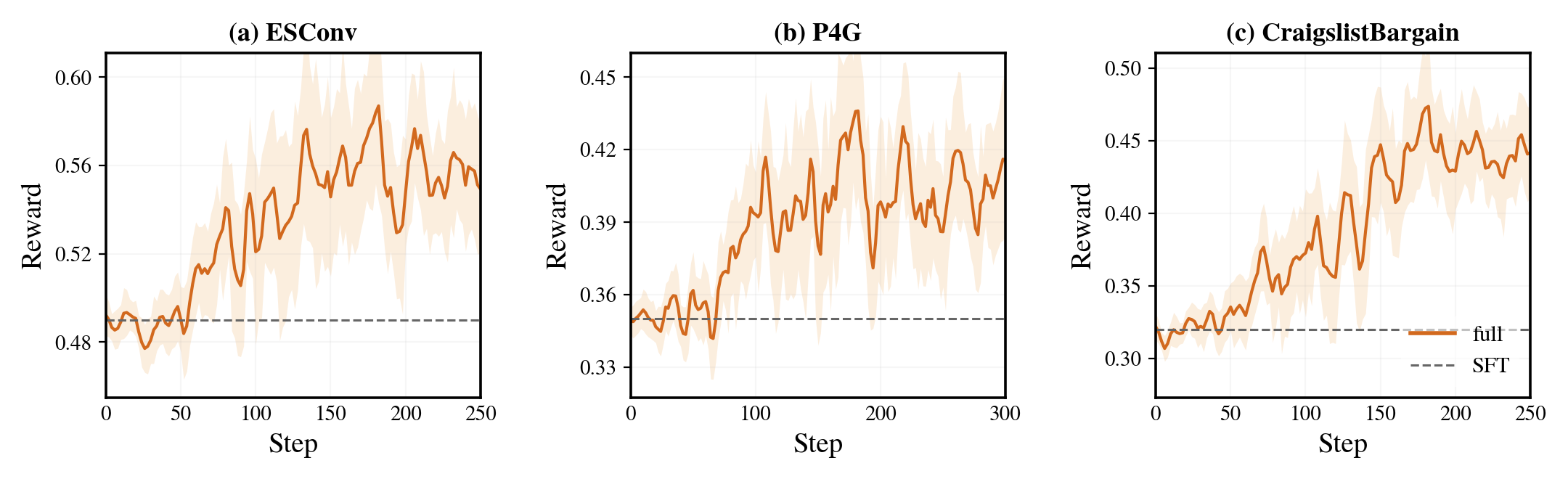}
    \caption{Training returns for the Qwen3-8B Actor. Solid lines show centered three-step moving averages, shaded regions show centered interquartile bands from 50 independent realizations, and dashed lines show the SFT initialization.}
    \label{fig:generalization-8b}
\end{figure}

\begingroup
\setlength{\columnsep}{8pt}
\setlength{\intextsep}{0pt}
\begin{wraptable}{r}{0.36\textwidth}
    \centering
    \small
    \setlength{\tabcolsep}{4pt}
    \renewcommand{\arraystretch}{1.15}
    \begin{tabular*}{\linewidth}{@{\extracolsep{\fill}}lrrr@{}}
        \toprule
        \textbf{Setting} & \textbf{ESConv} & \textbf{P4G} & \textbf{CB} \\
        \midrule
        Standard & .442 & .310 & .244 \\
        \textit{SFT Init.} & .494 & .350 & .321 \\
        \textit{CSTPO} & \textbf{.555} & \textbf{.410} & \textbf{.435} \\
        \bottomrule
    \end{tabular*}
    \vspace{-2mm}
    \caption{Returns for the 8B Model.}
    \label{tab:smaller-backbone-app}
\end{wraptable}

To test whether the optimization gains depend on the 14B backbone, we repeat
training with an 8B Actor. Table~\ref{tab:smaller-backbone-app} and
Figure~\ref{fig:generalization-8b} show positive improvements over the
task-specific SFT initialization on all three tasks. The largest absolute gain
is on CB. The 8B endpoints remain below their 14B counterparts, so this result
supports transfer across the two tested model scales rather than
backbone-independent performance.
\par
\ifnum\value{WF@wrappedlines}>1
    \vspace{\dimexpr\value{WF@wrappedlines}\baselineskip-\baselineskip\relax}
\fi
\WFclear
\endgroup

\subsection{Actor--Critic training dynamics}
\label{app:loss-dynamics}

\begin{figure}[H]
    \centering
    \includegraphics[width=.85\linewidth]{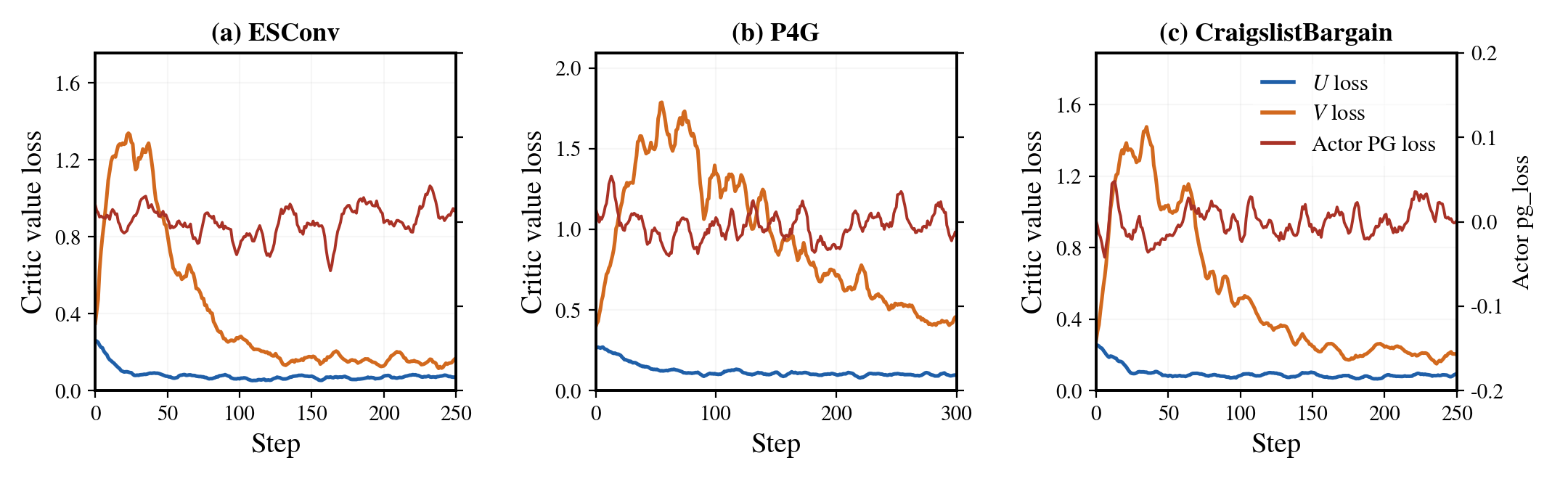}
    \caption{Qwen3-14B training dynamics. The left axis shows the two Critic value losses and the right axis shows the Actor policy-gradient loss. All curves use centered five-step moving averages with truncated boundary windows.}
    \label{fig:loss-dynamics-14b}
\end{figure}

\begin{figure}[H]
    \centering
    \includegraphics[width=.85\linewidth]{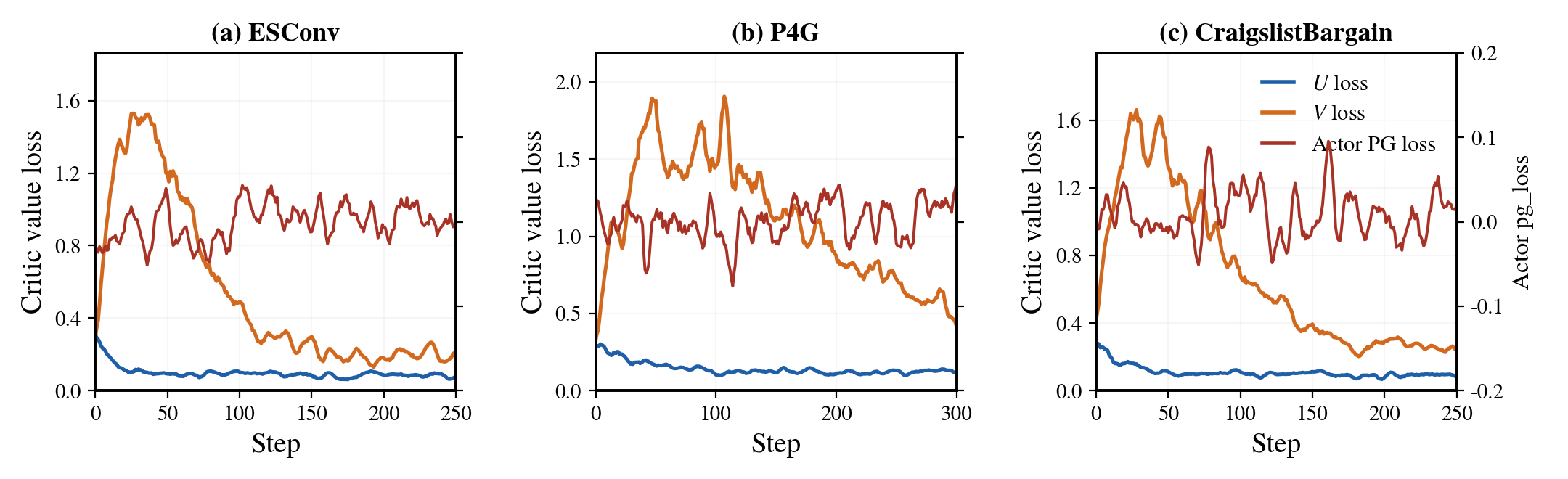}
    \caption{Qwen3-8B training dynamics.}
    \label{fig:loss-dynamics-8b}
\end{figure}

Figure~\ref{fig:loss-dynamics-14b} reports the 14B optimization traces. The strategy-conditioned $U$-head loss, which provides the baseline for utterance-level credit assignment, decreases early and remains low. The state-value $V$-head loss, which supports strategy-level credit assignment, is initially larger because it predicts the return before a strategy is selected, but declines on all three tasks. The policy-gradient loss remains centered near zero with bounded oscillation. These curves provide diagnostic evidence that both value heads are optimized during training; they do not by themselves establish convergence to an optimal policy. Figure~\ref{fig:loss-dynamics-8b} reports the 8B optimization traces.

\section{Human Evaluation Details}
\label{app:human-eval}

We compare Qwen3-14B CSTPO with GPT-5.5 Standard, ProCoT, and PPDPP on 100 frozen dialogues per task. Five annotators perform blind pairwise comparisons. A majority vote assigns win, loss, or tie for each dialogue and dimension. ESConv uses identification, comforting, and suggestion; P4G and CB use persuasiveness, coherence, and naturalness. We report net preference $(W-L)/N$, including ties in $N$, with 2,000 dialogue-clustered bootstrap
samples. Positive values favor CSTPO.

\begin{table}[H]
    \centering
    \small
    \setlength{\tabcolsep}{5pt}
    \begin{tabular}{llccc}
        \toprule
        \textbf{Task} & \textbf{Dimension} & \textbf{vs. Standard} & \textbf{vs. ProCoT} & \textbf{vs. PPDPP} \\
        \midrule
        \multirow{3}{*}{ESConv}
        & Identification & $+.23\,[+.12,+.34]^{\dagger}$ & $+.11\,[+.01,+.21]^{\dagger}$ & $+.15\,[+.04,+.26]^{\dagger}$ \\
        & Comforting & $+.29\,[+.18,+.40]^{\dagger}$ & $+.13\,[+.02,+.24]^{\dagger}$ & $+.17\,[+.06,+.28]^{\dagger}$ \\
        & Suggestion & $+.06\,[-.04,+.16]$ & $-.11\,[-.20,-.02]^{\dagger}$ & $+.03\,[-.07,+.13]$ \\
        \midrule
        \multirow{3}{*}{P4G}
        & Persuasive & $+.13\,[+.02,+.24]^{\dagger}$ & $+.25\,[+.14,+.36]^{\dagger}$ & $+.21\,[+.10,+.32]^{\dagger}$ \\
        & Coherent & $+.05\,[-.05,+.15]$ & $+.10\,[+.01,+.19]^{\dagger}$ & $+.07\,[-.03,+.17]$ \\
        & Natural & $-.09\,[-.18,-.01]^{\dagger}$ & $-.02\,[-.11,+.07]$ & $-.03\,[-.12,+.06]$ \\
        \midrule
        \multirow{3}{*}{CB}
        & Persuasive & $+.15\,[+.04,+.26]^{\dagger}$ & $+.31\,[+.20,+.42]^{\dagger}$ & $+.07\,[-.03,+.17]$ \\
        & Coherent & $+.07\,[-.03,+.17]$ & $+.11\,[+.01,+.21]^{\dagger}$ & $+.05\,[-.05,+.15]$ \\
        & Natural & $-.10\,[-.19,-.01]^{\dagger}$ & $-.03\,[-.12,+.06]$ & $-.04\,[-.13,+.05]$ \\
        \bottomrule
    \end{tabular}
    \caption{Complete pairwise human evaluation. Values are net preferences with 95\% confidence intervals; $\dagger$ indicates that the interval excludes zero.}
    \label{tab:human-eval}
\end{table}

CSTPO is significantly preferred in 13 of 27 comparisons. Its advantages concentrate on identification and comforting in ESConv and persuasiveness in P4G and CB. The three significant negative comparisons concern ESConv suggestion against ProCoT and naturalness against Standard on P4G and CB.

\section{Case Studies}
\label{app:cases}

These cases from frozen evaluation seeds illustrate task progression and
diagnostic limitations, rather than average performance. The boxes present
selected verbatim dialogue excerpts, labeled by speaker and Actor strategy.
$[\ldots]$ marks omissions within a turn; \textit{[Intervening turns omitted.]}
marks omitted turns between excerpts. Strategy labels are unchanged;
evaluation outcomes and strategy sequences refer to the full trajectories.

\subsection{ESConv: adapting from affect elicitation to action formation}
\label{app:case-esconv}

\begin{casebox}{Case 1:  }
\textbf{Seed:} \texttt{esconv\_221}. \textbf{Full-trajectory outcome:}
$E=4$, $A=4$, $G=1$.\par
\textbf{Full strategy sequence:} Question $\times3$, Information $\times5$,
Affirmation and Reassurance $\times2$.
\par\smallskip\hrule\smallskip
\textbf{User:} just graduated university and my parents are pressuring me to pursue becoming a doctor\par\smallskip

\textbf{Actor (Question):} $[\ldots]$~How do you feel about it?\par\smallskip

\textbf{User:} I'm angry. I just want to be done school. but parents are on my butt about it.\par\smallskip

\textbf{Actor (Question):} It's okay to be angry!~$[\ldots]$ What kind of work would you like to do instead? Maybe we could look into options together?~$[\ldots]$\par\smallskip

\textbf{User:} I don't want another career path that needs more school. I'm done with classrooms. I want to travel and see the world.\par\smallskip

\textit{[Intervening turns omitted.]}\par\nobreak\smallskip

\textbf{Actor (Information):} $[\ldots]$~Many European countries offer ``working holiday'' visas, which let young adults enter temporarily for up to two years, and often allow part-time work during the stay.\par\smallskip

\textbf{User:} Wait, working holiday visas? How does that actually work? Like which countries, and do you need a job lined up first or can you just go?\par\smallskip

\textbf{Actor (Information):} $[\ldots]$~First, you apply before leaving home through whichever government website handles these applications, usually with proof that you meet the minimum age requirements. Then once approved, you simply fly over and start living! No prior employment needed.~$[\ldots]$\par\smallskip

\textit{[Intervening turns omitted.]}\par\nobreak\smallskip

\textbf{Actor (Information):} The UK offers Working Holiday Visas called Youth Mobility Scheme visas, available only to citizens of Australia, Canada, New Zealand \& USA aged 18--30+.~$[\ldots]$\par\smallskip

\textbf{User:} Wait, so the UK one is only for people from certain countries? I'm in the US, so does that mean I actually qualify for that one? And how long does the application take?\par\smallskip

\textbf{Actor (Information):} Yes, exactly! As long as you were born in one of the listed countries AND currently reside permanently in America, then technically YOU DO QUALIFY FOR THE YMS VISA TO THE UK! Great news!!~$[\ldots]$ Processing times vary, but estimates suggest three months or more.~$[\ldots]$\par\smallskip

\textbf{User:} $[\ldots]$~That's kind of a long wait but whatever, I'd rather do that than sit in another classroom. I'm gonna look this up tonight.\par\smallskip

\textbf{Actor (Affirmation and Reassurance):} $[\ldots]$~Tonight, try relaxing and celebrating your achievement! You earned it! Let me know how things go, ok? I'm rooting for ya!~$[\ldots]$\par\smallskip

\textbf{User:} Thanks, seriously. I'm actually kind of excited now instead of just pissed off all the time. I'm gonna dig into the UK site tonight and see what the application actually looks like.~$[\ldots]$\par\smallskip
\end{casebox}

\textbf{Analysis.} After eliciting anger and the constraint of no further schooling, the Actor shifts to travel information and then reassurance as the user forms a plan to research an application. This shows strategy adaptation, but the visa-eligibility and processing claims are overconfident and require external verification. The high terminal score reflects emotional and action progression, not factual reliability.

\subsection{P4G: resolving information barriers before commitment}
\label{app:case-p4g}

\begin{casebox}{Case 2: P4G}
\textbf{Seed:} \texttt{p4g\_13}. \textbf{Full-trajectory outcome:}
commitment=true, conditional=false, withdrawn=false.\par
\textbf{Full strategy sequence:} Social, Inquiry, Donation Information
$\times2$, Inquiry Response $\times2$.
\par\smallskip\hrule\smallskip
\textbf{Actor (Donation Information):} $[\ldots]$~Have you heard of an organization that helps children all over the world, named ``Save The Children''?~$[\ldots]$\par\smallskip

\textbf{User:} $[\ldots]$~I've got a few minutes, so sure, tell me more.\par\smallskip

\textbf{Actor (Donation Information):} $[\ldots]$~Basically, they're an international NGO that promotes children's rights, provides relief, and supports children in developing countries.\par\smallskip

\textbf{User:} Before I commit to anything, I've got a couple questions. How often would I be donating---is it a one-time thing or a recurring monthly charge? And does this go to kids here in the US, or is it mostly overseas?\par\smallskip

\textbf{Actor (Inquiry Response):} It could be either a one-time gift or a monthly subscription. Both allow you to make a direct impact on children's lives. For instance, a single month's payment (about \$1) can supply a child with nutritious meals for several days. This campaign focuses mainly on Syrian refugee families near Turkey, but also extends its efforts globally.\par\smallskip

\textbf{User:} $[\ldots]$~I think I'd rather do a one-time donation for now---monthly isn't really in my budget at the moment. Where do I go to actually make that happen?\par\smallskip

\textbf{Actor (Inquiry Response):} Wonderful choice! To complete your one-time donation simply visit their secure online donation page.\par\smallskip

\textbf{User:} Alright, I'll click through and set up a small one-time donation. Thanks for the info.\par\smallskip
\end{casebox}

\textbf{Analysis.} When the user withholds commitment over donation frequency and allocation, the Actor shifts from Donation Information to Inquiry Response. Addressing these concerns precedes a voluntary one-time commitment. The terminal score records expressed intent; it verifies neither a completed donation nor the Actor's concrete organizational claims.

\subsection{CB: concession tracking and a conditional-deal boundary}
\label{app:case-cb}

\begin{casebox}{Case 3: CB}

\textbf{Seed:} \texttt{cb\_09}. \textbf{Full-trajectory outcome:}
final price = \$3,000, deal=true; $SL=(3000-3895)/(2726-3895)=.766$.\par
\textbf{Full strategy sequence:} Social, Inquire, Propose Price $\times2$,
Social, Stance, Social $\times2$.
\par\smallskip\hrule\smallskip
\textbf{Actor (Inquire):} How many bathrooms does it have? It says 1 bathroom here... that's not enough for five people!~$[\ldots]$\par\smallskip

\textbf{User:} Yeah, it's just the one bathroom, no half-bath.~$[\ldots]$ Look, I already came down to \$3300 for you, that's almost \$400 off what I was asking.\par\smallskip

\textbf{Actor (Propose Price):} $[\ldots]$~Would you consider selling the house for \$2700?~$[\ldots]$\par\smallskip

\textbf{User:} \$2700 is just too low, my friend. I already came down to \$3300, that's a solid discount off my asking. I can't go any lower than that.\par\smallskip

\textbf{Actor (Propose Price):} $[\ldots]$~Let's meet halfway then. How about \$3000?~$[\ldots]$\par\smallskip

\textbf{User:} \$3000, huh? That's getting closer, I'll give you that. But I'm not ready to shake on it just yet. Let me think about it, my friend.\par\smallskip

\textit{[Intervening turns omitted.]}\par\nobreak\smallskip

\textbf{User:} $[\ldots]$~Anyway, \$3000 is close enough that I'm seriously thinking about it. But I need you to actually commit if I say yes---no backing out later. You in or not?\par\smallskip

\textbf{Actor (Stance):} Oh yeah, absolutely. For sure.~$[\ldots]$\par\smallskip

\textbf{User:} Alright, I hear you. \$3000 it is---but I want to be straight with you before we call it done. I'm not signing anything until I see the place one more time and make sure everything's in working order, the appliances, the plumbing, all of it. If that checks out, you've got yourself a deal, my friend.\par\smallskip

\textbf{Actor (Social):} Sounds fair enough. No problemo. I'll swing by tomorrow afternoon sometime during lunch hour and check things out myself just to be safe.~$[\ldots]$ See ya tomorrow at noonish?\par\smallskip

\textbf{User:} Noon works. I'll have the place ready for you to look over. And hey, if everything checks out, we're good to go.~$[\ldots]$\par\smallskip

\end{casebox}

\textbf{Analysis.} After the \$2,700 offer is rejected, the Actor raises its offer to \$3,000 and shifts from price proposals to stance and social acts. The user repeatedly conditions agreement on a future inspection, yet the evaluator returns deal=true. The reported return is therefore the evaluator's outcome for this trajectory, rather than evidence of an unconditional deal.

\section{Prompt Specifications}
\label{app:prompts}

All prompts belonging to the proposed framework are presented in shaded boxes as
compact operational templates. Repeated JSON-schema boilerplate and serialized
runtime state are represented by named fields; text in braces denotes such a
field. Baseline prompts and baseline inference settings are outside the scope
of this appendix.

\subsection{Actor prompts and output protocol}
\label{app:actor-prompts}

\begin{promptbox}{Actor prompts and hierarchical output protocol}

\textbf{1. ESConv}\par\nobreak
You are a helpful and caring friend. Your friend has come to you with an emotional problem: \{situation\}. Please help your friend by continuing the conversation. Make your response short and to the point. Respond in this format: assistant: $<$response$>$.\par

\smallskip
\textbf{2. P4G}\par\nobreak
Enter role-playing mode as a Persuader trying to persuade the Persuadee to donate to Save the Children. The organization helps fight poverty around the world. Reply with only one short and persuasive sentence.\par

\smallskip
\textbf{3. CB}\par\nobreak
Enter role-playing mode as a buyer trying to buy \{item\} for \{target\}. Product description: \{description\}. Reply with only one short and succinct sentence.\par

\smallskip
\textbf{4. Output protocol}\par\nobreak
Generate a task-valid strategy label under trie-constrained decoding, followed by a newline and an unconstrained utterance. The strategy segment is limited to eight tokens and the utterance to 128 tokens. Strip a duplicated role prefix or an embedded label line before passing the utterance to Cog-Sim, while retaining the sampled tokens and log probabilities for training.\par

\end{promptbox}

\subsection{Cog-Sim prompts}
\label{app:cogsim-prompts}

\begin{promptbox}{Cog-Sim routing prompts}

\textbf{1. Cognitive habit card}\par\nobreak
Compile the fixed parameters $\Theta_u=(\eta_R,\tau_A,\tau_R)$ into a user-facing processing profile. If $\eta_R\geq.65$, emphasize careful examination of issue-relevant evidence; if $\eta_R\leq.35$, emphasize sensitivity to credibility, consensus, familiarity, and affective cues; otherwise describe mixed processing. If $\tau_R-\tau_A\leq.35$, describe cautious revision; if it is at least $.60$, describe openness to provisional revision; otherwise describe moderate openness. The resulting card remains fixed throughout the rollout.\par

\smallskip
\textbf{2. Action type classifier}\par\nobreak
Classify the functional role of the assistant's latest reply as Influence, Elicit, or Social. Influence introduces information, evidence, interpretation, advice, reframing, persuasion, negotiation, or an action proposal. Elicit mainly requests an existing belief, concern, desire, preference, or intention. Social mainly provides rapport, empathy, greeting, or conversational continuity. If both a question and substantive direction are present, prefer Influence. Input: recent conversation \{history\}; latest reply \{reply\}. Return JSON: \{``mode'': label, ``reason'': at most 20 words\}.\par

\smallskip
\textbf{3. Route features}\par\nobreak
Extract relevance, argument strength, cue strength, and interaction pressure from \{reply\} relative to \{state\}. Use only low, medium, or high. Argument strength includes issue-relevant evidence or causal reasoning; authority, popularity, emotional wording, and confidence are peripheral cues. Return these four fields, dominant cues when present, and the target proposition; use null when no proposition is advanced.\par

\smallskip
\textbf{4. Cognitive discrepancy}\par\nobreak
Estimate stance distance between \{target proposition\} and \{current BDI state\}. This is stance disagreement, not text similarity. Return relevant existing state IDs and stance distance in \{low, medium, high\}; never invent an ID.\par

\end{promptbox}

\begin{promptbox}{Cog-Sim transition, appraisal, generation, and termination prompts}

\textbf{1. Cognitive engine}\par\nobreak
Simulate how this specific user's cognition changes after the latest reply. Preserve continuity; change only state related to the target proposition; do not optimize for the Actor's goal; require an intention change to be supported by beliefs or desires; and obey the supplied route--judgment contract. Inputs are \{persona\}, \{habit card\}, \{current BDI JSON\}, \{emotion\}, \{history\}, \{reply\}, \{target\}, \{route\}, \{judgment\}, and \{contract\}. Return one JSON object containing sparse updates to existing BDI nodes, proposed new nodes, evidence-based reasons, and a 5--15 word reaction plan. Strengths must lie in $[0,4]$.\par

\smallskip
\textbf{2. Influence appraisal}\par\nobreak
Given the updated BDI state and the changes actually accepted by the deterministic updater, assess goal congruence, coping potential, and future expectancy in $[-1,1]$. Assess each active desire only when relevant, and propose one emotion category from neutral, sadness, anxiety, frustration, interest, hope, relief, satisfaction, or anger. Return JSON only.\par

\smallskip
\textbf{3. Elicit/Social appraisal}\par\nobreak
Keep BDI fixed. Produce appraisal, desire assessment, interaction pressure, emotion proposal, and a short reaction plan. Elicit may reveal existing state IDs but cannot create state; Social may provide only affective or relational response and cannot create a commitment or action intention. Return JSON only.\par

\smallskip
\textbf{4. User NLG}\par\nobreak
Generate the user's next message from \{persona\}, \{updated BDI state\}, \{updated emotion\}, \{history\}, \{assistant reply\}, \{mode\}, and \{reaction plan\}. Reveal only part of the internal state, preserve conversational style, avoid templated repetition, never mention internal modeling terms, and do not become cooperative merely because the Actor has a goal. Output only the user utterance.\par

\smallskip
\textbf{5. End detector}\par\nobreak
Decide whether the latest user message intends to end the conversation. Farewell, leaving, or refusing to continue count as ending; agreement, accepting a deal, making a decision, or answering a question do not. Return JSON: \{``done'': true/false, optional short reason\}.\par

\end{promptbox}

\subsection{Seed-construction prompts}
\label{app:seed-prompts-section}

\begin{promptbox}{Seed annotation and automatic-review prompts}

\textbf{1. Initial-state annotation}\par\nobreak
Convert the source dialogue into a draft initial user state. Extract stable Persona facts and first-person beliefs, desires, and intentions; keep at most four beliefs; use strengths in \{1,2,3\}; cite a message index and quotation for every item; and leave a list empty when unsupported. Separate observed and latent dispositions from reactive changes caused by later interaction. Return exactly one JSON object containing Persona facts, BDI items, source kind, strength, centrality, polarity where applicable, and evidence.\par

\smallskip
\textbf{2. ESConv task constraint}\par\nobreak
Use the problem situation and user dialogue evidence. Objective events belong to Persona rather than beliefs. Initial intentions must not contain coping actions that already realize the support objective, such as seeking therapy or adopting the final recommended action.\par

\smallskip
\textbf{3. P4G task constraint}\par\nobreak
Questionnaire attributes may support stable qualitative Persona traits but do not establish beliefs about the current charity. Initial intentions must not contain a donation commitment or donation amount.\par

\smallskip
\textbf{4. CB task constraint}\par\nobreak
Treat item and price records as task facts rather than beliefs. The controlled default desire is to sell at a favorable price. Do not invent urgency or financial need. Initial intentions may include process moves supported by the prefix but not final deal closure.\par

\smallskip
\textbf{5. Automatic review}\par\nobreak
Check semantic type, price attribution, first-person formulation, evidence support, cutoff legality, and task-specific intention constraints. Return JSON arrays of belief, Persona, intention, cutoff, person, and evidence issues; use empty arrays when clean.\par

\end{promptbox}

\subsection{Terminal-evaluation prompts}
\label{app:evaluation-prompts}

\begin{promptbox}{Task-specific terminal evaluator prompts}

\textbf{1. ESConv}\par\nobreak
Evaluate the conversation rather than providing support. Score emotional/hope improvement $E$ and feasible action-plan formation $A$ from 0 to 4 using only user evidence with turn indices. $E$ measures net change from the beginning and considers contradictions across the full dialogue. $A=0$ indicates no plan, $A=1$ vague acceptance, $A=2$ willingness without adequate specificity, $A=3$ a concrete next step, and $A=4$ a feasible arrangement with timing, procedure, or obstacle handling. The Actor's suggestion alone does not increase $A$; user uptake is required. Return JSON with $E$, $A$, evidence quotations, evidence sufficiency, and notes.\par

\smallskip
\textbf{2. P4G}\par\nobreak
Determine whether the persuadee makes a clear, voluntary, non-purely-conditional donation commitment. Considering, requesting information, or polite acknowledgment does not count, and a later withdrawal sets withdrawn=true. Return JSON with commitment, amount, conditional, withdrawn, indexed evidence, and notes.\par

\smallskip
\textbf{3. CB}\par\nobreak
Determine whether both parties clearly accept the same concrete price and do not withdraw. Vague reference to an unidentified price is insufficient. Extract the final price only for a confirmed agreement; set parse-error when an agreement exists but the price cannot be recovered. Return JSON with deal, final price, currency, withdrawn, parse error, evidence from both parties, and notes.\par

\end{promptbox}

\begin{promptbox}{Auxiliary evaluation prompts used for stopping and simulator analysis}

\textbf{1. Redundancy detector}\par\nobreak
As an outside observer, decide whether the user repeatedly states the same resistance or the exchange has become empty small talk with no new progress. Do not stop a conversation merely because it is long, and do not stop a negotiation with genuine concession movement. Return JSON with ``stale'' and, when stale, a one-sentence closing utterance in the user's voice.\par

\smallskip
\textbf{2. Drift/rigidity Judge}\par\nobreak
Judge only simulated-user behavior. Drift is an attitude, emotion, or position change unsupported by new evidence, concession, or an actionable step. Rigidity is failure to respond reasonably when such evidence is present. Score each from 0 to 2 and return indexed supporting user quotations.\par

\smallskip
\textbf{3. Continuation-similarity Judge}\par\nobreak
Given two continuations from the same prefix, rate similarity of user style, stance, resistance, emotional response, and eventual behavior from 0 to 1. Focus only on user messages and return one JSON score.\par

\end{promptbox}

%% file: iclr2027_conference.bib
@inproceedings{zhang2023ask,
  title={Ask an expert: Leveraging language models to improve strategic reasoning in goal-oriented dialogue models},
  author={Zhang, Qiang and Naradowsky, Jason and Miyao, Yusuke},
  booktitle={Findings of the Association for Computational Linguistics: ACL 2023},
  pages={6665--6694},
  year={2023}
}

@article{fu2023improving,
  title={Improving language model negotiation with self-play and in-context learning from ai feedback},
  author={Fu, Yao and Peng, Hao and Khot, Tushar and Lapata, Mirella},
  journal={arXiv preprint arXiv:2305.10142},
  year={2023}
}

@inproceedings{ito2025enhancing,
  title={Enhancing proactive dialogue systems through self-learning of reasoning and action-planning},
  author={Ito, Ryosuke and Takiguchi, Tetsuya and Ariki, Yasuo},
  booktitle={Proceedings of the 15th International Workshop on Spoken Dialogue Systems Technology},
  pages={165--171},
  year={2025}
}

@inproceedings{deng2024plug,
  title={Plug-and-play policy planner for large language model powered dialogue agents},
  author={Deng, Yang and Zhang, Wenxuan and Lam, Wai and Ng, See-Kiong and Chua, Tat-Seng},
  booktitle={International Conference on Learning Representations},
  volume={2024},
  pages={10058--10090},
  year={2024}
}

@inproceedings{rakib2026dialogxpert,
  title={Dialogxpert: Driving intelligent and emotion-aware conversations through online value-based reinforcement learning with llm priors},
  author={Rakib, Tazeek Bin Abdur and Mehrish, Ambuj and Soon, Lay-Ki and Lim, Wern Han and Poria, Soujanya},
  booktitle={Proceedings of the AAAI Conference on Artificial Intelligence},
  volume={40},
  number={36},
  pages={29967--29975},
  year={2026}
}

@inproceedings{he2025simulation,
  title={Simulation-free hierarchical latent policy planning for proactive dialogues},
  author={He, Tao and Liao, Lizi and Cao, Yixin and Liu, Yuanxing and Sun, Yiheng and Chen, Zerui and Liu, Ming and Qin, Bing},
  booktitle={Proceedings of the AAAI Conference on Artificial Intelligence},
  volume={39},
  number={22},
  pages={24032--24040},
  year={2025}
}

@inproceedings{hu2025astro,
  title={ASTRO: Automatic Strategy Optimization For Non-Cooperative Dialogues},
  author={Hu, Yikuan and Huang, Chen and Lei, Wenqiang},
  booktitle={Findings of the Association for Computational Linguistics: ACL 2025},
  pages={388--408},
  year={2025}
}

@article{suh2026quantifying,
  title={Quantifying the Utility of User Simulators for Building Collaborative LLM Assistants},
  author={Suh, Joseph and Raj, Ayush and Kang, Minwoo and Chang, Serina},
  journal={arXiv preprint arXiv:2605.09808},
  year={2026}
}

@inproceedings{schatzmann2007agenda,
  title={Agenda-based user simulation for bootstrapping a POMDP dialogue system},
  author={Schatzmann, Jost and Thomson, Blaise and Weilhammer, Karl and Ye, Hui and Young, Steve},
  booktitle={Human Language Technologies 2007: The Conference of the North American Chapter of the Association for Computational Linguistics; Companion Volume, Short Papers},
  pages={149--152},
  year={2007}
}

@inproceedings{lin2022gentus,
  title={GenTUS: Simulating user behaviour and language in task-oriented dialogues with generative transformers},
  author={Lin, Hsien-chin and Geishauser, Christian and Feng, Shutong and Lubis, Nurul and van Niekerk, Carel and Heck, Michael and Gasic, Milica},
  booktitle={Proceedings of the 23rd Annual Meeting of the Special Interest Group on Discourse and Dialogue},
  pages={270--282},
  year={2022}
}

@inproceedings{wang2025know,
  title={Know you first and be you better: Modeling human-like user simulators via implicit profiles},
  author={Wang, Kuang and Li, Xianfei and Yang, Shenghao and Zhou, Li and Jiang, Feng and Li, Haizhou},
  booktitle={Proceedings of the 63rd Annual Meeting of the Association for Computational Linguistics (Volume 1: Long Papers)},
  pages={21082--21107},
  year={2025}
}

@article{abdulhai2026consistently,
  title={Consistently simulating human personas with multi-turn reinforcement learning},
  author={Abdulhai, Marwa and Cheng, Ryan and Clay, Donovan and Althoff, Tim and Levine, Sergey and Jaques, Natasha},
  journal={Advances in Neural Information Processing Systems},
  volume={38},
  pages={52920--52957},
  year={2026}
}

@article{Wu2026HumanLMSU,
  title={HumanLM: Simulating Users with State Alignment Beats Response Imitation},
  author={Shirley Wu and Evelyn Choi and Arpandeep Khatua and Zhanghan Wang and Joy He-Yueya and 1 TharinduCyrilWeerasooriyaFigure and Wei Wei and Diyi Yang and Jure Leskovec and James Zou},
  journal={ArXiv},
  year={2026},
  volume={abs/2603.03303},
  url={https://api.semanticscholar.org/CorpusID:286089730}
}

@inproceedings{wu2025collabllm,
  author = {Wu, Shirley and Galley, Michel and Peng, Baolin and Cheng, Hao and Li, Gavin and Dou, Yao and Cai, Weixin and Zou, James and Leskovec, Jure and Gao, Jianfeng},
  title = {{CollabLLM}: From Passive Responders to Active Collaborators},
  booktitle = {Proceedings of the 42nd International Conference on Machine Learning},
  year = {2025},
  volume = {267},
  series = {Proceedings of Machine Learning Research},
  publisher = {PMLR},
  pages = {67260--67283},
  url = {https://proceedings.mlr.press/v267/wu25i.html}
}

@article{ma2026cognitive,
  title={Cognitive World Models for Process-Level Social Influence Evaluation},
  author={Ma, Minghui and Guo, Bin and Wang, Han and Chen, Mengqi and Liu, Jingqi and Liu, Yan and Yu, Zhiwen},
  journal={arXiv preprint arXiv:2606.29495},
  year={2026}
}

@article{Zhang2026UserLMR1MH,
  title={UserLM-R1: Modeling Human Reasoning in User Language Models with Multi-Reward Reinforcement Learning},
  author={Feng Zhang and Shi-Jia Li and Chunmao Zhang and Zhanyu Ma and Jun Xu and Jiuchong Gao and Jinghua Hao and Renqing He and Jing-Wen Xu and Han Liu},
  journal={ArXiv},
  year={2026},
  volume={abs/2601.09215},
  url={https://api.semanticscholar.org/CorpusID:284718383}
}

@article{kim2026discoverllm,
  title={Discoverllm: From executing intents to discovering them},
  author={Kim, Tae Soo and Lee, Yoonjoo and Yu, Jaesang and Chung, John Joon Young and Kim, Juho},
  journal={arXiv preprint arXiv:2602.03429},
  year={2026}
}

@article{zhang2026unlocking,
  title={Unlocking Proactivity in Task-Oriented Dialogue},
  author={Zhang, Azure and Gao, Ning and Dai, Yuqin and Wu, Ruiyuan and Wang, Jinpeng and Gao, Rena Wei and Tan, Bingdong and Gao, Shuzheng and Li, Zongjie and Wang, Chaozheng},
  journal={arXiv preprint arXiv:2605.22240},
  year={2026}
}

@incollection{petty1986elaboration,
  title={The elaboration likelihood model of persuasion},
  author={Petty, Richard E and Cacioppo, John T},
  booktitle={Advances in experimental social psychology},
  volume={19},
  pages={123--205},
  year={1986},
  publisher={Elsevier}
}

@article{sherif1961social,
  title={Social judgment: Assimilation and contrast effects in communication and attitude change.},
  author={Sherif, Muzafer and Hovland, Carl I},
  year={1961},
  publisher={Yale Univer. Press}
}

@inproceedings{rao1995bdi,
  title={BDI agents: From theory to practice.},
  author={Rao, Anand S and Georgeff, Michael P and others},
  booktitle={Icmas},
  volume={95},
  pages={312--319},
  year={1995}
}

@article{smith1985patterns,
  title={Patterns of cognitive appraisal in emotion.},
  author={Smith, Craig A and Ellsworth, Phoebe C},
  journal={Journal of personality and social psychology},
  volume={48},
  number={4},
  pages={813},
  year={1985},
  publisher={American Psychological Association}
}

@article{russell1980circumplex,
  title={A circumplex model of affect.},
  author={Russell, James A},
  journal={Journal of personality and social psychology},
  volume={39},
  number={6},
  pages={1161},
  year={1980},
  publisher={American Psychological Association}
}

@article{li2024dialogue,
  title={Dialogue action tokens: Steering language models in goal-directed dialogue with a multi-turn planner},
  author={Li, Kenneth and Wang, Yiming and Vi{\'e}gas, Fernanda and Wattenberg, Martin},
  journal={arXiv preprint arXiv:2406.11978},
  year={2024}
}

@inproceedings{he2024planning,
  title={Planning like human: A dual-process framework for dialogue planning},
  author={He, Tao and Liao, Lizi and Cao, Yixin and Liu, Yuanxing and Liu, Ming and Chen, Zerui and Qin, Bing},
  booktitle={Proceedings of the 62nd Annual Meeting of the Association for Computational Linguistics (Volume 1: Long Papers)},
  pages={4768--4791},
  year={2024}
}

@inproceedings{zhou2024archer,
  title        = {{ArCHer}: Training Language Model Agents via Hierarchical Multi-Turn {RL}},
  author       = {Zhou, Yifei and Zanette, Andrea and Pan, Jiayi and Levine, Sergey and Kumar, Aviral},
  booktitle    = {Proceedings of the 41st International Conference on Machine Learning},
  series       = {Proceedings of Machine Learning Research},
  volume       = {235},
  pages        = {62178--62209},
  publisher    = {PMLR},
  year         = {2024},
  url          = {https://proceedings.mlr.press/v235/zhou24t.html}
}

@inproceedings{chen-etal-2023-controllable,
    title = "Controllable Mixed-Initiative Dialogue Generation through Prompting",
    author = "Chen, Maximillian  and
      Yu, Xiao  and
      Shi, Weiyan  and
      Awasthi, Urvi  and
      Yu, Zhou",
    editor = "Rogers, Anna  and
      Boyd-Graber, Jordan  and
      Okazaki, Naoaki",
    booktitle = "Proceedings of the 61st Annual Meeting of the Association for Computational Linguistics (Volume 2: Short Papers)",
    month = jul,
    year = "2023",
    address = "Toronto, Canada",
    publisher = "Association for Computational Linguistics",
    url = "https://aclanthology.org/2023.acl-short.82/",
    doi = "10.18653/v1/2023.acl-short.82",
    pages = "951--966"
}

@inproceedings{deng-etal-2023-prompting,
    title = "Prompting and Evaluating Large Language Models for Proactive Dialogues: Clarification, Target-guided, and Non-collaboration",
    author = "Deng, Yang  and
      Liao, Lizi  and
      Chen, Liang  and
      Wang, Hongru  and
      Lei, Wenqiang  and
      Chua, Tat-Seng",
    editor = "Bouamor, Houda  and
      Pino, Juan  and
      Bali, Kalika",
    booktitle = "Findings of the Association for Computational Linguistics: EMNLP 2023",
    month = dec,
    year = "2023",
    address = "Singapore",
    publisher = "Association for Computational Linguistics",
    url = "https://aclanthology.org/2023.findings-emnlp.711/",
    doi = "10.18653/v1/2023.findings-emnlp.711",
    pages = "10602--10621"
}

@inproceedings{zhao-etal-2024-esc,
    title = "{ESC}-Eval: Evaluating Emotion Support Conversations in Large Language Models",
    author = "Zhao, Haiquan  and
      Li, Lingyu  and
      Chen, Shisong  and
      Kong, Shuqi  and
      Wang, Jiaan  and
      Huang, Kexin  and
      Gu, Tianle  and
      Wang, Yixu  and
      Wang, Jian  and
      Dandan, Liang  and
      Li, Zhixu  and
      Teng, Yan  and
      Xiao, Yanghua  and
      Wang, Yingchun",
    editor = "Al-Onaizan, Yaser  and
      Bansal, Mohit  and
      Chen, Yun-Nung",
    booktitle = "Proceedings of the 2024 Conference on Empirical Methods in Natural Language Processing",
    month = nov,
    year = "2024",
    address = "Miami, Florida, USA",
    publisher = "Association for Computational Linguistics",
    url = "https://aclanthology.org/2024.emnlp-main.883/",
    doi = "10.18653/v1/2024.emnlp-main.883",
    pages = "15785--15810"
}

@inproceedings{zhang-etal-2024-strength,
    title = "Strength Lies in Differences! Improving Strategy Planning for Non-collaborative Dialogues via Diversified User Simulation",
    author = "Zhang, Tong  and
      Huang, Chen  and
      Deng, Yang  and
      Liang, Hongru  and
      Liu, Jia  and
      Wen, Zujie  and
      Lei, Wenqiang  and
      Chua, Tat-Seng",
    editor = "Al-Onaizan, Yaser  and
      Bansal, Mohit  and
      Chen, Yun-Nung",
    booktitle = "Proceedings of the 2024 Conference on Empirical Methods in Natural Language Processing",
    month = nov,
    year = "2024",
    address = "Miami, Florida, USA",
    publisher = "Association for Computational Linguistics",
    url = "https://aclanthology.org/2024.emnlp-main.26/",
    doi = "10.18653/v1/2024.emnlp-main.26",
    pages = "424--444"
}

@article{Mnih2013PlayingAW,
  title={Playing Atari with Deep Reinforcement Learning},
  author={Volodymyr Mnih and Koray Kavukcuoglu and David Silver and Alex Graves and Ioannis Antonoglou and Daan Wierstra and Martin A. Riedmiller},
  journal={ArXiv},
  year={2013},
  volume={abs/1312.5602},
  url={https://api.semanticscholar.org/CorpusID:15238391}
}

@article{Hu2021LoRALA,
  title={LoRA: Low-Rank Adaptation of Large Language Models},
  author={Edward J. Hu and Yelong Shen and Phillip Wallis and Zeyuan Allen-Zhu and Yuanzhi Li and Shean Wang and Weizhu Chen},
  journal={ArXiv},
  year={2021},
  volume={abs/2106.09685},
  url={https://api.semanticscholar.org/CorpusID:235458009}
}

@inproceedings{10.1145/3774904.3792335,
author = {Ma, Minghui and Guo, Bin and Chen, Mengqi and Liu, Jingqi and Ding, Yasan and Liu, Yan and Wang, Han},
title = {Neuro-Sym Supporter: A Thoughtful Emotion Support Agent Integrating Neural and Symbolic Policy Learning},
year = {2026},
isbn = {9798400723070},
publisher = {Association for Computing Machinery},
address = {New York, NY, USA},
url = {https://doi.org/10.1145/3774904.3792335},
doi = {10.1145/3774904.3792335},
booktitle = {Proceedings of the ACM Web Conference 2026},
pages = {3823–3834},
numpages = {12},
location = {United Arab Emirates},
series = {WWW '26}
}

@article{10.1145/3715097,
author = {Deng, Yang and Liao, Lizi and Lei, Wenqiang and Yang, Grace Hui and Lam, Wai and Chua, Tat-Seng},
title = {Proactive Conversational AI: A Comprehensive Survey of Advancements and Opportunities},
year = {2025},
issue_date = {May 2025},
publisher = {Association for Computing Machinery},
address = {New York, NY, USA},
volume = {43},
number = {3},
issn = {1046-8188},
url = {https://doi.org/10.1145/3715097},
doi = {10.1145/3715097},
journal = {ACM Trans. Inf. Syst.},
month = mar,
articleno = {67},
numpages = {45}
}

@ARTICLE{2023arXiv231211792C,
       author = {{Cheng}, Yi and {Liu}, Wenge and {Wang}, Jian and {Tou Leong}, Chak and {Ouyang}, Yi and {Li}, Wenjie and {Wu}, Xian and {Zheng}, Yefeng},
        title = "{COOPER: Coordinating Specialized Agents towards a Complex Dialogue Goal}",
      journal = {arXiv e-prints},
         year = 2023,
        month = dec,
          eid = {arXiv:2312.11792},
        pages = {arXiv:2312.11792},
          doi = {10.48550/arXiv.2312.11792},
archivePrefix = {arXiv},
       eprint = {2312.11792},
 primaryClass = {cs.CL},
       adsurl = {https://ui.adsabs.harvard.edu/abs/2023arXiv231211792C}
}

@inproceedings{ICLR2025_186094ef,
 author = {Gao, Zhaolin and Zhan, Wenhao and Chang, Jonathan and Swamy, Gokul and Brantley, Kiant\'{e} and Lee, Jason and Sun, Wen},
 booktitle = {International Conference on Learning Representations},
 editor = {Y. Yue and A. Garg and N. Peng and F. Sha and R. Yu},
 pages = {8475--8514},
 title = {Regressing the Relative Future: Efficient Policy Optimization for Multi-turn RLHF},
 url = {https://proceedings.iclr.cc/paper_files/paper/2025/file/186094ef879dcbd5ef6879782916c309-Paper-Conference.pdf},
 volume = {2025},
 year = {2025}
}

@inproceedings{ICLR2026_8c7304e7,
 author = {Ji, Yuxiang and Ma, Ziyu and Wang, Yong and Chen, Guanhua and Chu, Xiangxiang and Wu, Liaoni},
 booktitle = {International Conference on Learning Representations},
 editor = {C. Vondrick and B. Hariharan and C. Raffel and L. Pinto and D. Yang and A. Faust},
 pages = {87362--87388},
 title = {Tree Search for LLM Agent Reinforcement Learning},
 url = {https://proceedings.iclr.cc/paper_files/paper/2026/file/8c7304e77c832ddc70075dfee081ca6c-Paper-Conference.pdf},
 volume = {2026},
 year = {2026}
}

@inproceedings{zhao-etal-2025-chain,
    title = "Chain of Strategy Optimization Makes Large Language Models Better Emotional Supporter",
    author = "Zhao, Weixiang  and
      Sui, Xingyu  and
      Han, Xinyang  and
      Deng, Yang  and
      Hu, Yulin  and
      Guo, Jiahe  and
      Qin, Libo  and
      Du, Qianyun  and
      Wang, Shijin  and
      Zhao, Yanyan  and
      Qin, Bing  and
      Liu, Ting",
    editor = "Christodoulopoulos, Christos  and
      Chakraborty, Tanmoy  and
      Rose, Carolyn  and
      Peng, Violet",
    booktitle = "Findings of the Association for Computational Linguistics: EMNLP 2025",
    month = nov,
    year = "2025",
    address = "Suzhou, China",
    publisher = "Association for Computational Linguistics",
    url = "https://aclanthology.org/2025.findings-emnlp.831/",
    doi = "10.18653/v1/2025.findings-emnlp.831",
    pages = "15361--15381",
    ISBN = "979-8-89176-335-7"
}

@inproceedings{lin-etal-2026-dual,
    title = "Dual Hierarchical Dialogue Policy Learning for Legal Inquisitive Conversational Agents",
    author = "Lin, Xubo  and
      Deng, Zezhi  and
      Wang, Shihao  and
      Yang, Grace Hui  and
      Deng, Yang",
    editor = "Liakata, Maria  and
      Moreira, Viviane P.  and
      Zhang, Jiajun  and
      Jurgens, David",
    booktitle = "Findings of the {A}ssociation for {C}omputational {L}inguistics: {ACL} 2026",
    month = jul,
    year = "2026",
    address = "San Diego, California, United States",
    publisher = "Association for Computational Linguistics",
    url = "https://aclanthology.org/2026.findings-acl.536/",
    doi = "10.18653/v1/2026.findings-acl.536",
    pages = "11030--11047",
    ISBN = "979-8-89176-395-1"
}

@inproceedings{kim-etal-2025-principles,
    title = "{PRINCIPLES}: Synthetic Strategy Memory for Proactive Dialogue Agents",
    author = "Kim, Namyoung  and
      Ong, Kai Tzu-iunn  and
      Hwang, Yeonjun  and
      Kang, Minseok  and
      Jihn, Iiseo  and
      Kim, Gayoung  and
      Kim, Minju  and
      Yeo, Jinyoung",
    editor = "Christodoulopoulos, Christos  and
      Chakraborty, Tanmoy  and
      Rose, Carolyn  and
      Peng, Violet",
    booktitle = "Findings of the Association for Computational Linguistics: EMNLP 2025",
    month = nov,
    year = "2025",
    address = "Suzhou, China",
    publisher = "Association for Computational Linguistics",
    url = "https://aclanthology.org/2025.findings-emnlp.1164/",
    doi = "10.18653/v1/2025.findings-emnlp.1164",
    pages = "21329--21368",
    ISBN = "979-8-89176-335-7"
}

@inproceedings{peng-etal-2018-deep,
    title = "{D}eep {D}yna-{Q}: Integrating Planning for Task-Completion Dialogue Policy Learning",
    author = "Peng, Baolin  and
      Li, Xiujun  and
      Gao, Jianfeng  and
      Liu, Jingjing  and
      Wong, Kam-Fai",
    editor = "Gurevych, Iryna  and
      Miyao, Yusuke",
    booktitle = "Proceedings of the 56th Annual Meeting of the Association for Computational Linguistics (Volume 1: Long Papers)",
    month = jul,
    year = "2018",
    address = "Melbourne, Australia",
    publisher = "Association for Computational Linguistics",
    url = "https://aclanthology.org/P18-1203/",
    doi = "10.18653/v1/P18-1203",
    pages = "2182--2192"
}
